%% file: main.tex
\documentclass{article} 
\usepackage{iclr2027_conference,times}

\input{math_commands.tex}

\usepackage{hyperref}
\hypersetup{colorlinks=true,linkcolor=blue,citecolor=blue,urlcolor=blue,pdfauthor={}}
\usepackage{url}
\usepackage{graphicx}
\usepackage{wrapfig}
\usepackage{booktabs}
\usepackage{multirow}
\usepackage{caption}
\usepackage{xcolor}         
\usepackage{wrapfig}
\usepackage{caption}
\usepackage{enumitem}
\newcommand{\best}[1]{\textbf{#1}}
\title{Rethinking Asymmetry in Self-Supervised 3D Geometry Learning}

\author{
\hspace*{-\tabcolsep}%
\parbox{\textwidth}{\centering
\vspace*{1em}%
\textbf{Gyeongjin Kang}$^{1}$ \quad
\textbf{Freeman Cheng}$^{2}$ \quad
\textbf{Botao Ye}$^{3}$ \quad
\textbf{Weijie Lyu}$^{2}$ \\[0.2em]
\textbf{Ming-Hsuan Yang}$^{2}$ \quad
\textbf{Eunbyung Park}$^{1}$ \\[0.6em]
{\normalfont
$^{1}$Yonsei University \quad
$^{2}$UC Merced \quad
$^{3}$ETH Zurich} \\[0.8em]
{\small
\href{https://gynjn.github.io/rethinking-asymmetry/}{\raisebox{-0.12em}{\includegraphics[height=0.82em]{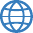}}\hspace{0.28em}Website}
\quad
\href{https://github.com/Gynjn/rethinking-asymmetry}{\raisebox{-0.12em}{\includegraphics[height=0.82em]{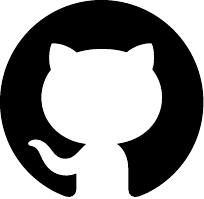}}\hspace{0.28em}Code}}
}
}

\iclrfinalcopy 
\begin{document}

\maketitle
\lhead{Preprint}

\begin{figure}[h]
    \centering
    \vspace{-0.5cm}
    \includegraphics[width=\textwidth]{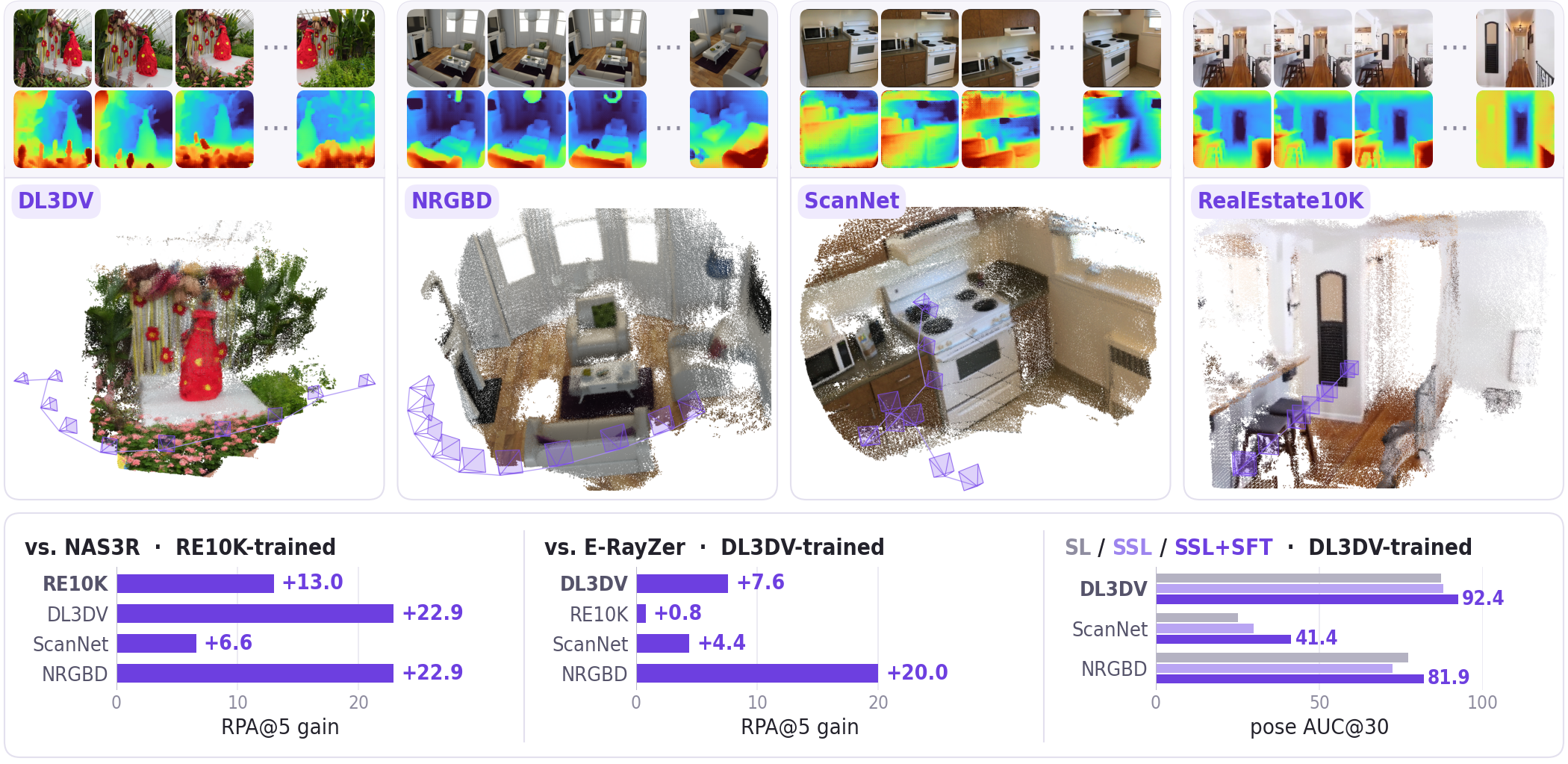}
    \caption{From unposed images alone, our self-supervised model recovers camera poses and 3D geometry in a single forward pass (top), generalizes to unseen datasets, improves pose accuracy over baseline models (bottom left, middle), and provides an effective initialization for supervised fine-tuning (SSL+SFT), compared to supervised training only (bottom right).}
    \label{fig:teaser}
\end{figure}

\begin{abstract}
\textcolor{black}{Self-supervised learning relies on asymmetry between what a model observes and what it must explain. Recent methods learn 3D geometry from unposed images by predicting cameras and scene structure, then rendering held-out views. To avoid trivial solutions, they impose this asymmetry in the input space, restricting how views interact through separate branches or masked attention. We revisit this design with geometric accuracy, rather than rendering quality, as the main objective. When a held-out view lies between context views, the context already constrains the geometry it observes. We therefore impose the asymmetry in the 3D representation space, excluding held-out views only from the reconstructed scene while a shared backbone uses all views for correspondence. This improves depth and pose estimation over input-space designs. We further exploit view density: sparse and dense observations depict the same scene but provide different amounts of correspondence. Aligning sparse-context features with dense-context features supplies a geometric training signal alongside rendering, without a pretrained geometric teacher. We additionally align features with a frozen vision foundation model, which brings further gains. None of these signals requires geometric annotation, and the resulting model improves geometry estimation over prior self-supervised methods. Controlled experiments isolate the contribution of each component, and self-supervised pretraining also improves subsequent supervised training.}
\end{abstract}

\section{Introduction} 

Self-supervised learning has driven the progress in representation learning. Because it requires no manual labels, models can be pretrained on as much data as we can collect, and their performance continues to improve as data, model size, and compute scale~\citep{kaplan2020scaling}. This approach has succeeded in language~\citep{radford2018gpt, devlin2019bert}, vision~\citep{caron2021dino, he2022mae}, and embodied control~\citep{rhoda2026webvideo}. In contrast, many 3D foundation models are trained to reproduce geometric annotations recovered by optimization pipelines such as COLMAP~\citep{schoenberger2016sfm}, which are costly to run and hard to scale. Recent work~\citep{jiang2025rayzer, zhao2026erayzer, huang2026nas} removes this dependence by making novel-view synthesis (NVS) the training objective, predicting cameras and scene structure from raw images or a video and rendering held-out views using its own predictions. These methods differ in scene representation and how they share information across views, but evaluations focus largely on synthesis quality. In self-supervised learning, however, the pretext task is only a means, and a method is judged by what it learns. We therefore treat view synthesis as the pretext and geometry as the outcome, and ask which designs make geometry learning more accurate and scalable.

We begin with the scene representation. Since the model learns about the scene only through what it renders, the form of this representation determines how closely the learned structure reflects the actual 3D geometry. Latent NVS methods~\citep{jiang2025rayzer, mitchel2026xfactor, prestel2026rayder} render held-out views from an implicit scene code, so geometry is never made explicit. They achieve strong rendering quality, but can do so without recovering structure that corresponds to physical 3D space, leaving geometry and camera motion only weakly constrained by the synthesis loss. We instead use an explicit representation, 3D Gaussians~\citep{kerbl20233dgs} placed in physical space, so that every rendered pixel depends directly on the predicted depth and camera pose.

\begin{wrapfigure}{r}{0.50\textwidth}
  \begin{center}
  \vspace{-2mm}   
    \includegraphics[width=0.92\linewidth, trim=0 0 0 10]{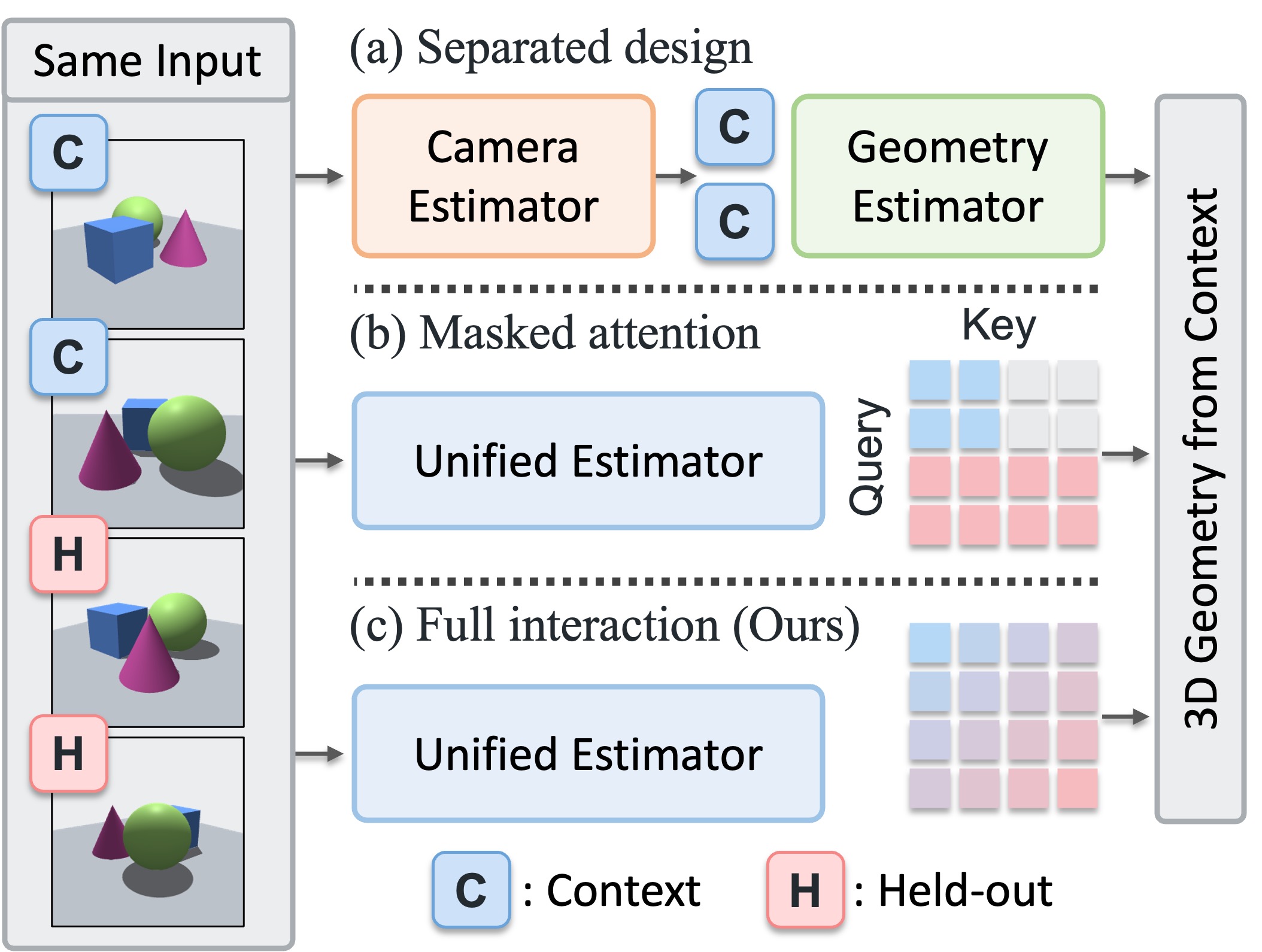}
\caption{All designs build the 3D scene from context views (C). (a) Separate backbones and (b) masked attention restrict interaction between context and held-out views (H). (c) Our design allows full cross-view attention while excluding held-out views from the 3D representation.}
    \label{fig:inf_profile}
    \end{center}
    \vspace{-5mm}   
\end{wrapfigure}

\textbf{Where should the asymmetry live?} A self-supervised objective needs an asymmetry between what the model sees and what it must explain. Otherwise, trivial solutions satisfy it. For example, in masked image modeling~\citep{he2022mae, yang2026pixio}, the image itself is the only place to create this asymmetry, so masking parts of it directly defines the task. In contrast, masking a view in multi-view 3D does not create the same asymmetry. When a held-out view lies between context views (interpolative NVS setting), the 3D scene reconstructed from the context views is the same scene that all views together describe. Removing the held-out view therefore hides nothing the model must recover. It only restricts the correspondence that would have helped recover that scene more accurately. Existing methods nevertheless impose the asymmetry on the images the network observes\footnote{This prevents held-out views from leaking into the reconstruction, which matters when rendering quality is the measure of interest. Geometry could also become biased toward the held-out views, but we find that richer view context in fact yields better geometric estimation (Sec.~\ref{app:asym}).}. E-RayZer~\citep{zhao2026erayzer} splits camera and geometry prediction into two backbones so that the geometry branch sees only context views, and NAS3R~\citep{huang2026nas} masks view interaction between the context and held-out sets, so context views attend only among themselves. Both keep held-out views out of the reconstruction, but in doing so also withhold the correspondence those views could provide.

We instead place the asymmetry in the 3D representation space. A shared backbone lets all views exchange information when predicting depth and pose, so every view contributes correspondence. We keep the context and held-out split of prior work but apply it when constructing the 3D scene, so that a held-out view is rendered only from Gaussians predicted for context views.  This is equivalent to masking the 3D representation rather than the images. Every view thus informs correspondence estimation, while the self-reconstruction path is blocked.

\textbf{Asymmetric feature alignment.} View density provides a second source of asymmetry. The same model predicts more accurate depth and pose when it sees more views of a scene, so a model given fewer views is a weaker version of itself. Such gaps within a single model have been used to bootstrap representations without any external teacher~\citep{jiang2025sra, chefer2026selfflow}. SelfEvo~\citep{huang2026selfevo} applies this idea to 3D, refining a model pretrained with 3D supervision~\citep{wang2025vggt, wang2026pi} by using the geometry and camera predictions of a dense-context teacher as pseudo labels for a sparse-context student. We instead build this asymmetry into self-supervised training, optimizing it jointly with the rendering objective without any pretrained geometric teacher. In this setting the dense-context predictions are themselves unreliable, and matching them can propagate their errors. We therefore align the features that produce the geometry rather than the geometry itself. The largest gains occur in single-view depth estimation, which the rendering objective does not supervise directly.

\textbf{Semantics as a geometric prior.} Both signals above are grounded in rendered pixels. The rendering loss supervises the model directly, and asymmetric context alignment does so indirectly through features shaped by rendering. Pixel agreement, however, says little about which pixels belong together, such as which regions form the same surface or object. Such high-level structure often helps constrain geometry where photometric cues are weak, such as textureless or repetitive regions. Vision foundation models capture it well, and aligning with their features has proven effective in generative modeling~\citep{yu2024repa, wu2026reg} and supervised perception tasks~\citep{yang2024da, knaebel2026dinoroom}. We use such alignment as a semantic prior for self-supervised learning, aligning our backbone features with those of a frozen vision foundation model~\citep{simeoni2025dinov3, yang2026pixio},  obtaining consistent gains in geometry estimation at modest additional cost.

We combine these components into a single self-supervised geometry framework and evaluate each through controlled experiments. Our contributions are as follows.

\begin{itemize}[leftmargin=*,nosep]
\item \textbf{Asymmetry in the 3D representation.} We show that imposing the asymmetry on the reconstructed scene rather than on the input images prevents trivial solutions while letting a shared backbone use every view for correspondence, improving depth and pose estimation.
\item \textbf{Asymmetric feature alignment.} We align sparse-context features with dense-context features jointly with the rendering objective, improving geometry without a pretrained geometric teacher.
\item \textbf{Semantics as a geometric prior.} Aligning intermediate features with a frozen vision foundation model adds structural guidance beyond pixel agreement, improving depth and pose estimation without geometric annotations.
\end{itemize}

\section{Related Work}
\vspace{-2mm}
\textbf{Self-supervised learning with novel-view synthesis.} Feed-forward 3D models recover geometry directly from a set of images, but they are trained against camera and depth targets that an offline pipeline must produce first~\citep{wang2024dust3r, wang2025vggt, wang2026vggt, wang2026pi, lin2025da3}. A separate line of work has instead learned geometry from images alone. Early methods predicted depth and ego-motion from video and used warping between neighboring frames as supervision~\citep{zhou2017sfmlearner, godard2019digging, bian2019scdepth, lai2021videoautoencoder}, which ties learning to short baselines and pairwise constraints. RUST~\citep{sajjadi2023rust} and RayZer~\citep{jiang2025rayzer} removed explicit geometry altogether, encoding a scene into a latent set~\citep{sajjadi2022srt} and rendering new views with a learned decoder. Several methods extend this latent formulation to dynamic data and predictive (extrapolative) objectives~\citep{chen2026wildrayzer, mitchel2026xfactor, prestel2026rayder}. Rendering in latent space minimizes inductive bias and lets the model learn freely from data, but it also leaves the recovered geometry only loosely tied to physical structure.

Another line of work grounds the scene in 3D space and renders it by volume rendering or rasterization, jointly predicting poses and a NeRF~\citep{mildenhall2021nerf} or Gaussian scene~\citep{kerbl20233dgs} under photometric losses~\citep{chen2023dbarf, fu2023mononerf, kang2025selfsplat}. Recently, E-RayZer~\citep{zhao2026erayzer} and NAS3R~\citep{huang2026nas} scaled explicit self-supervised reconstruction to large transformers and data, yielding representations with stronger geometric grounding. To prevent shortcuts and information leakage, both restrict how information moves between the context and held-out views, E-RayZer by separating camera and geometry prediction into two backbones, NAS3R by masking view interaction in its single backbone. The restriction keeps the objective from collapsing, but at the cost of rich correspondence. We instead investigate where this restriction should be placed, and find that the scene representation is a better location than the input.

\textbf{Self-supervised representation alignment.} Aligning intermediate features to an additional target has become a common way to improve representation learning beyond what the task loss provides. In generative modeling, REPA~\citep{yu2024repa} and follow-up work~\citep{wu2026reg, shin2026pixelrepa} align diffusion transformer features~\citep{peebles2023dit} with a frozen vision foundation model~\citep{oquab2023dinov2}, accelerating convergence and improving sample quality. Perception models use the same idea, aligning task features with such encoders~\citep{yang2024da, knaebel2026dinoroom} or distilling several of them into one~\citep{ranzinger2024radio, zhu2026eupe}.

On the other hand, recent work derives the target from the model itself, comparing its own representations across different inputs or layers~\citep{jiang2025sra, chefer2026selfflow}. SelfEvo~\citep{huang2026selfevo} brings this to 3D, distilling a dense-context teacher into a sparse-context student, though it starts from a supervised model and aligns the two branches through their predicted depth. We exploit the same sparse-dense gap without 3D-supervised initialization, aligning features jointly with the main rendering objective during training. We also align them with a frozen vision encoder, providing complementary internal and external signals without geometric annotations.

\section{Approach}
\label{sec:approach}
\subsection{Preliminaries}

\textbf{Setup.} Given $N$ unposed images $\mathcal{V}=\{I_i\}_{i=1}^{N}$ of a static scene, we partition them into disjoint context and target sets, $\mathcal{C}$ and $\mathcal{T}$. A model $f_\theta$ predicts a scene representation $\mathcal{S}$ and camera poses $\hat{\pi}_i=(\hat{T}_i,\hat{K}_i)$, where $\hat{T}_i\in SE(3)$ is the (camera-to-world) extrinsics and $\hat{K}_i$ the intrinsics.
A renderer $g$, either a learned neural decoder or a differentiable rasterizer, then synthesizes each target view:
\begin{equation}
(\mathcal{S},\{\hat{\pi}_i\})=f_\theta(\mathcal{V}),\quad
\hat{I}_j=g(\mathcal{S},\hat{\pi}_j),\quad
\mathcal{L}_{\text{nvs}}=\sum_{j\in\mathcal{T}}\ell(\hat{I}_j,I_j).
\label{eq:nvs}
\end{equation}
Cameras are predicted relative to a reference view, and since no metric supervision is available, the reconstruction is recovered only up to scale. Estimating the pose of a target view requires observing its image, so target images must enter the network. If the representation predicted from a target view also enters $\mathcal{S}$, however, the model can render that view and satisfy Eq.~\ref{eq:nvs} without recovering a scene shared across views. Therefore, target views must be excluded from the scene used to render them.

\textbf{Implicit vs.\ explicit representation.} Implicit methods encode the context views into latent scene tokens $Z$, and render a target view with a learned decoder $g_\phi$ that queries $Z$ using an encoding of the target camera. Since $Z$ is not tied to physical 3D space, a low photometric loss need not reflect a consistent 3D scene, and the decoder can rely on interpolating appearance between nearby context views. Explicit methods instead represent the scene in 3D space, for example as 3D Gaussians. For each pixel $u$ of view $i$, an output head predicts a ray depth $d_{i,u}$ along with the Gaussian attributes. The Gaussian center is obtained by lifting the depth through the predicted camera,
\begin{equation}
\mu_{i,u} = \hat{T}_i\, d_{i,u}\, \frac{\hat{K}_i^{-1} u}{\lVert \hat{K}_i^{-1} u \rVert},
\qquad 
\mathcal{S}=\textstyle\bigcup_{i\in\mathcal{C}}\mathcal{G}_{i},
\qquad
\mathcal{G}_{i} = \textstyle\bigcup_{u}\mathcal{G}_{i,u},
\label{eq:lift}
\end{equation}
where $u$ is in homogeneous coordinates and $\mathcal{G}_{i}$ is a set of Gaussians predicted from view $i$, and $\mathcal{S}$ collects them over all context views. Rendering it with a differentiable rasterizer $g$ places both depth and pose on the rendering path. A target view is then well explained only when predictions from different views agree in a shared 3D space, so photometric error becomes a proxy for multi-view geometric consistency.

\textbf{Where asymmetry is placed.} Existing methods keep target views out of the reconstructed scene by restricting how information flows between views. \emph{Separate branches}~\citep{zhao2026erayzer} predict cameras from all views but geometry from context views alone, using separate features without the interaction that a shared representation can provide. \emph{Masked attention}~\citep{huang2026nas} uses a single backbone with a view-level mask $A\in\{0,1\}^{N\times N}$, setting $A_{ik}=0$ for $i\in\mathcal{C}$ and $k\in\mathcal{T}$, so that context tokens never read from targets. This keeps a shared backbone but removes all context-to-target interaction. Both designs thus break the trivial solution by restricting which views the model can relate in the input space.

\begin{wrapfigure}{r}{0.65\textwidth}
  \begin{center}
  \vspace{-3mm}   
    \includegraphics[width=0.92\linewidth, trim=0 0 0 10]{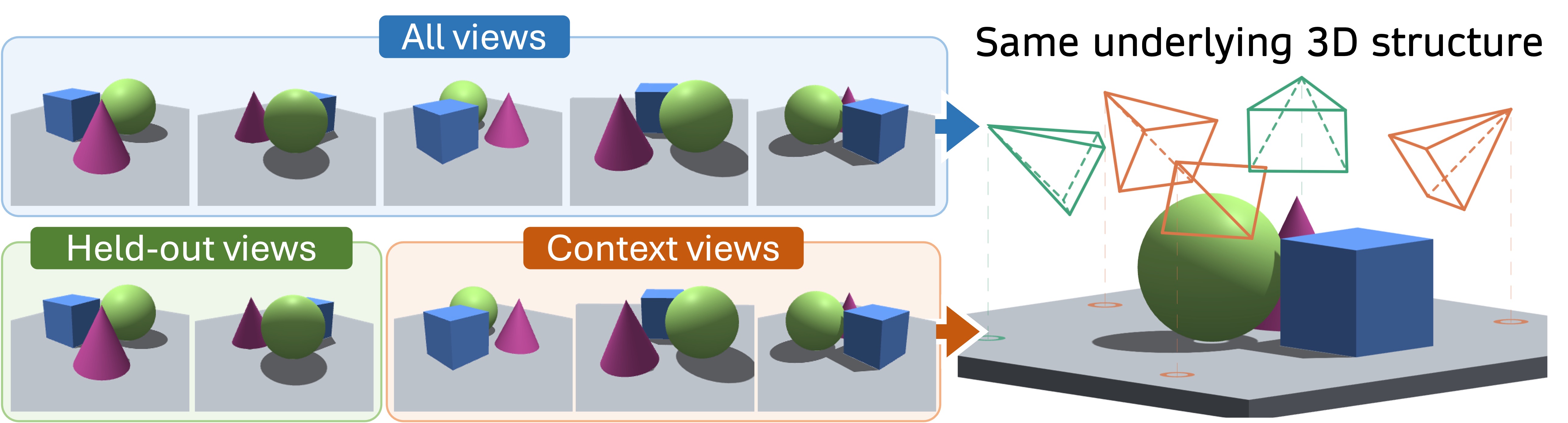}
    \caption{The context views alone determine the same scene as all captured views. All views are shown in blue, the context subset in orange, and held-out targets in green. Held-out views add correspondence, not new geometry.}
    \label{fig:same_scene}
    \end{center}
    \vspace{-5mm}   
\end{wrapfigure}

\subsection{Asymmetry in the 3D Representation}
\label{app:asym}
We now ask where the asymmetry should be placed when the goal is geometry rather than rendering. Prior designs keep target views out of view interaction so that the model cannot observe what it is asked to render. However, when a target view lies between context views, the context views already cover most of the geometry the target observes (Fig.~\ref{fig:same_scene}), so the target contributes correspondence rather than new scene content. The asymmetry therefore only needs to prevent a target view from being rendered from its own prediction. It need not restrict which views interact. We test this against existing placements, holding the representation, encoder, and training setup fixed.

\begin{figure}[!h]
    \vspace{-3mm}
    \centering
    \includegraphics[width=1.0\linewidth]{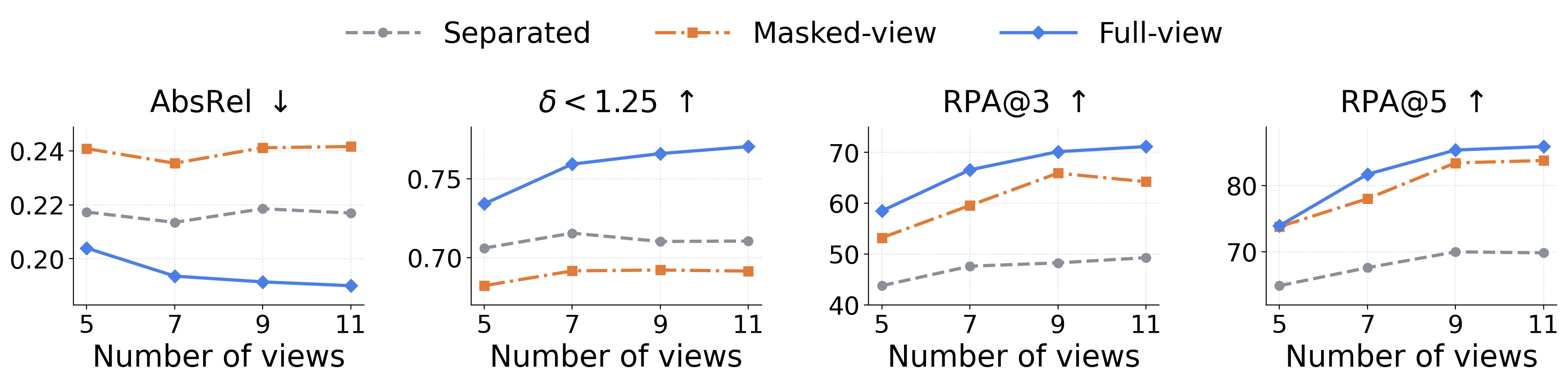}
\caption{Effect of full interaction on depth and pose estimation. Depth is compared with MoGe-2~\citep{wang2026moge2} pseudo ground truth, and pose with dataset cameras.}
\label{fig:input_asym}
    \vspace{-3mm}
\end{figure}

\textbf{Controlled comparison.} We compare three placements of asymmetry. All variants share a visual encoder~\citep{yang2026pixio} followed by a 12-layer transformer~\citep{vaswani2017attention} with alternating view-wise and global attention~\citep{wang2025vggt}. The full-view variant (ours) and the masked-attention variant predict cameras and Gaussians from this backbone. The separate-branch variant follows E-RayZer, predicting cameras from the backbone and Gaussians from a second 12-layer transformer, which gives it more capacity. All variants are trained on DL3DV~\citep{ling2024dl3dv} for 100K iterations with the same data, view sampling, and optimization settings, with the number of views varying across training batches. At evaluation, we fix the frame span at 48 frames, so fewer views provide sparser coverage of the same scene (Fig.~\ref{fig:input_asym}). Despite having less capacity than the separate-branch variant, the full-view variant achieves the most accurate depth and pose at every view count. Placing the asymmetry in the 3D representation rather than on interaction between views lets every view contribute correspondence while still blocking the self-reconstruction path.

\subsection{Beyond Rendering Signals}
\textbf{Learning from context asymmetry.} The preceding analysis shows that geometry estimates become less accurate as the view set becomes sparser. Dense and sparse predictions come from the same network on the same scene, so the gap reflects only what the extra views contribute, and this asymmetry can serve as a learning signal. SelfEvo~\citep{huang2026selfevo} turns this gap into supervision for a pretrained 3D foundation model~\citep{wang2025vggt, wang2026pi}, using the outputs of a dense-context teacher, maintained as an EMA of the model, as pseudo labels for a sparse-context student.

We build on this idea with one key difference. Our model is trained entirely through self-supervision, with no 3D supervision to anchor its predictions. In this setting, depth predicted from different view sets shares no common scale, and the dense-context depth can itself be unreliable. In our experiments, distilling it directly degrades geometry even with a scale-invariant loss. We therefore move the alignment from the outputs to the features that produce them. These features carry the cross-view correspondence that dense context provides, but are not tied to a particular scale or depth value, making them a more stable target than the outputs they produce. We match them with a cosine similarity loss, which constrains their direction but leaves their magnitude free. An EMA copy $f_{\bar{\theta}}$ processes the full set $\mathcal{V}$, and the online network $f_\theta$ processes a subset $\mathcal{V}_s\subset\mathcal{V}$. We align the patch features of each view, token by token, at the layers $\mathcal{K}$ that feed the output heads,
\begin{equation}
\mathcal{L}_{\text{align}} = -\frac{1}{|\mathcal{K}|\,|\mathcal{V}_s|\,P} \sum_{k \in \mathcal{K}} \sum_{i \in \mathcal{V}_s} \sum_{p=1}^{P} \cos\!\big(h_k(F^{(k)}_{s,i,p}),\, \operatorname{sg}[F^{(k)}_{i,p}]\big), \qquad \cos(a, b) = \frac{a^\top b}{\|a\|\,\|b\|},
\label{eq:align}
\end{equation}
where $F = f_{\bar{\theta}}(\mathcal{V})$ and $F_s = f_{\theta}(\mathcal{V}_s)$ are the dense and sparse features, $p$ indexes the $P$ patch tokens of each view, $h_k$ is a lightweight projection for layer $k$, $\operatorname{sg}[\cdot]$ is the stop-gradient, and $\bar{\theta}$ is an exponential moving average of $\theta$.

\begin{figure}[!h]
\vspace{-2mm}
\centering
\includegraphics[width=\linewidth]{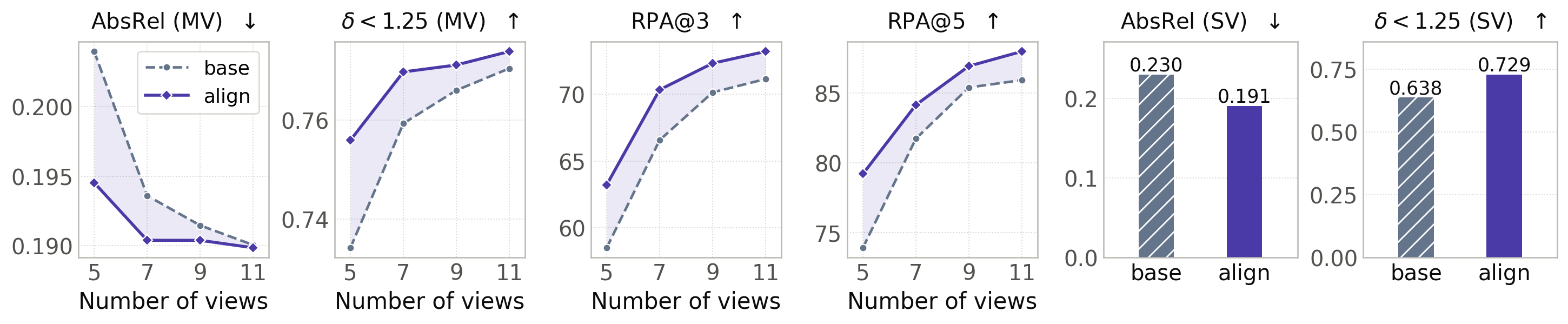}
\caption{Effect of asymmetric feature alignment on DL3DV dataset. The left four plots evaluate multi-view depth and camera pose; the right two evaluate depth from a single input image.}
\label{fig:selfalign}
\vspace{-2mm}
\end{figure}
We train the full-view model from Sec.~\ref{app:asym} with and without asymmetric context alignment, holding all other settings fixed. Alignment improves depth and pose at every view count, with larger gains when fewer views are available and the largest on single-view depth, the limiting case of sparse context where rendering provides no direct supervision (Fig.~\ref{fig:selfalign}).

\textbf{Semantic guidance.} The rendering loss supervises appearance at the pixel level, so the constraints it provides are local. In textureless or repetitive regions, photometric error barely changes with depth and says little about surface boundaries. Such distinctions are what semantic features encode, so we align our encoder features with those of a frozen vision foundation model~\citep{simeoni2025dinov3, yang2026pixio}.
For each view, the encoder features $F_i^{\text{enc}}$ are matched to the corresponding patch features of the frozen model $\Phi$ with a cosine similarity loss,
\begin{equation}
\mathcal{L}_{\text{sem}} = -\frac{1}{|\mathcal{V}|\,P} \sum_{i \in \mathcal{V}} \sum_{p=1}^{P} \cos\!\big(F^{\text{enc}}_{i,p},\, \Phi(I_i)_p\big).
\label{eq:sem}
\end{equation}
Since $\Phi$ is itself self-supervised, this adds no annotation. Applying the alignment at the encoder output, before the multi-view transformer, leaves the later layers free to specialize for geometry. Although the encoder is initialized from the same model, initialization alone does not retain this prior under the rendering loss, and the alignment preserves it throughout training.

\subsection{Training Objectives}
Our model is trained with three signals. The rendering loss defines the self-supervised task, asymmetric context alignment transfers dense-context features to sparse inputs, and semantic guidance injects a prior from a frozen foundation model. The total objective is
\begin{equation}
\mathcal{L}=\mathcal{L}_{\text{nvs}}+\lambda_{\text{align}}\mathcal{L}_{\text{align}}+\lambda_{\text{sem}}\mathcal{L}_{\text{sem}},
\label{eq:total}
\end{equation}
where $\mathcal{L}_{\text{nvs}}$ combines an L2 and a perceptual loss~\citep{chen2017photographic}, and the $\lambda$ terms balance the auxiliary losses against it. We set $\lambda_{\text{align}}=0.1$ and $\lambda_{\text{sem}}=0.2$. 
Since the dense-context features are only meaningful once rendering has shaped them, $\mathcal{L}_{\text{align}}$ is introduced at 40\% of training and ramped to full weight by 60\%.

\section{Experiments}
\label{sec:experiments}
\textbf{Implementation details.} Our model uses a pretrained Pixio (base) visual encoder~\citep{yang2026pixio} followed by a 12-layer multi-view transformer with an alternating attention pattern, a feature dimension of $768$, and a 16$\times$16 patch size. All experiments use this same backbone, which predicts both cameras and Gaussians. The camera head regresses each pose relative to the middle input view and predicts one set of intrinsics shared across the scene. A DPT head~\citep{ranftl2021dpt, wang2026vggt} predicts per-pixel Gaussian attributes, which are lifted to 3D with the predicted cameras. The scene is built from context-view Gaussians only, and target views are rendered at their predicted poses by differentiable rasterization~\citep{kerbl20233dgs}. Asymmetric context alignment uses the four transformer layers read by the DPT head, and semantic guidance uses the encoder output. Sparse subsets are formed by symmetrically dropping views on either side of the reference view, and the dense branch is an EMA copy of the online network with momentum 0.999.

\textbf{Datasets.} We train separate models on RealEstate10K (RE10K)~\citep{zhou2018stereo} and DL3DV~\citep{ling2024dl3dv}, with images resized to $256\times256$. ScanNetv2~\citep{dai2017scannetv2} and NRGBD~\citep{azinovic2022nrgbd} are used only for out-of-domain evaluation.

\textbf{Evaluation.} We use publicly released baseline checkpoints and follow their evaluation settings. On RE10K, we adopt the view indices of SPFSplat~\citep{huang2025spfsplat}, and on DL3DV, we follow E-RayZer~\citep{zhao2026erayzer} and sample 10 views across a 96-frame span. For out-of-domain evaluation, we sample 10 views at a fixed stride, spanning 64 frames on ScanNetv2 and 256 on NRGBD. We report camera pose and depth, the two geometric quantities of interest. Camera pose is measured by relative pose accuracy (RPA) at several angular thresholds and by AUC@30. Depth is measured by AbsRel and threshold accuracy $\delta < 1.25^{k}$ after independent per-view median alignment against ground-truth depth. Since DL3DV provides no ground-truth depth, our controlled analyses on DL3DV use MoGe-2~\citep{wang2026moge2} predictions as pseudo ground truth. Multi-view depth is predicted from the full input set, and single-view depth from each image alone.

\textbf{Depth conversion.} All depth metrics use planar $z$-depth. Our model and E-RayZer predict depth $r$ along each unit ray (Eq.~\ref{eq:lift}), which we convert as $z = r/\lVert \hat{K}^{-1}\mathbf{u} \rVert$ using the predicted intrinsics $\hat{K}$, so focal-length errors are reflected in the depth metrics. For NAS3R, we use the $z$-depth from its depth head. For SPFSplat, we transform its canonical point map into each view with the predicted pose and take the $z$ coordinate. 

\subsection{Results}
\begin{figure}[!h]
    \vspace{-4mm}
    \centering
    \includegraphics[width=1.0\linewidth]{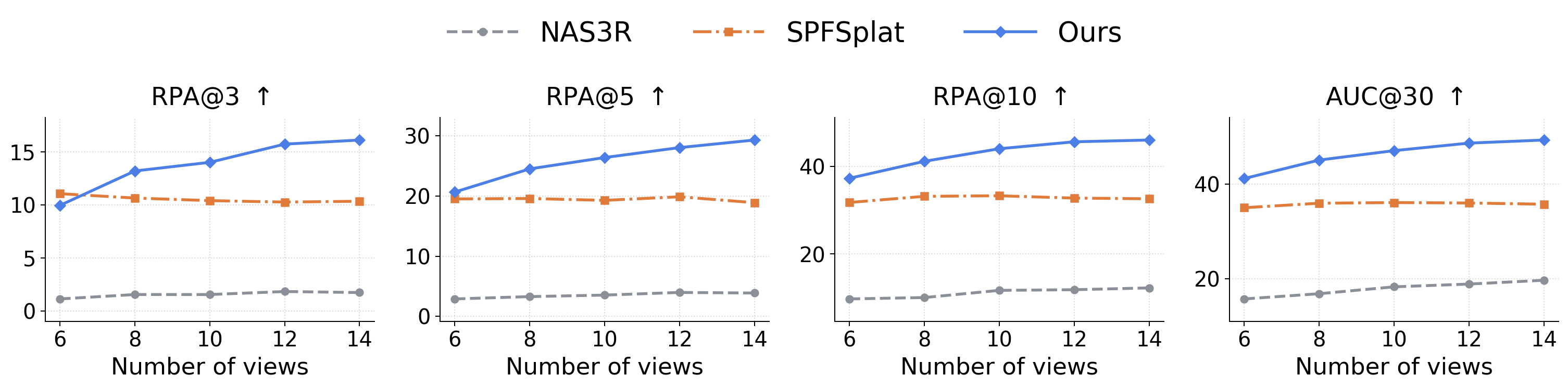}
\caption{Camera pose accuracy on DL3DV dataset as the number of input views increases.}
\label{fig:dl3dv_views}
    \vspace{-3mm}
\end{figure}

We compare with SPFSplat and NAS3R~\citep{huang2026nas}, which learn cameras and geometry from unposed images using self-supervised rendering signal. All three methods train on RE10K with varying numbers of input views. SPFSplat additionally initializes its backbone from MASt3R~\citep{leroy2024grounding}, which is trained with 3D supervision on multiple annotated datasets, and also takes ground-truth intrinsics as input. With 10 input views, our model achieves the highest pose accuracy at every reported threshold and competitive or better depth accuracy, despite using far fewer parameters. Its advantage grows on out-of-domain data (Tab.~\ref{tab:pose_re10k}). We further examine how each method scales with the number of input views on an out-of-domain dataset. Our model improves steadily on every pose metric as views are added (Fig.~\ref{fig:dl3dv_views}).

\begin{table}[!h]
  \centering
  \resizebox{\columnwidth}{!}{
  \begin{tabular}{lcccccccccccccc}
  \toprule
  \multirow{2}{*}{Method} & \multirow{2}{*}{3D prior} & \multirow{2}{*}{Params} & \multicolumn{2}{c}{RE10K (trained)} &
\multicolumn{2}{c}{DL3DV} & \multicolumn{4}{c}{ScanNetv2} & \multicolumn{4}{c}{NRGBD} \\
  \cmidrule(lr){4-5} \cmidrule(lr){6-7} \cmidrule(lr){8-11} \cmidrule(lr){12-15}
  & & & RPA@3$\uparrow$ & @5$\uparrow$ & RPA@3$\uparrow$ & @5$\uparrow$ & RPA@3$\uparrow$ & @5$\uparrow$ & AbsRel$\downarrow$ & $\delta < 1.25\uparrow$ & RPA@3$\uparrow$ & @5$\uparrow$ & AbsRel$\downarrow$
& $\delta < 1.25\uparrow$ \\
  \midrule
  SPFSplat & MASt3R & 629M & 64.2 & 78.1 & 10.5 & 19.3 & 2.0 & 5.7 & \best{0.223} & 0.757 & 3.2 & 14.1 & 0.156 & 0.803 \\
  NAS3R    & -- & 1.22B & 47.4 & 67.2 & 1.5 & 3.5 & 0.1 & 0.5 & 0.381 & 0.604 & 3.7 & 7.2 & 0.214 & 0.630 \\
  Ours     & -- & 183M & \best{65.7} & \best{80.2} & \best{14.2} & \best{26.4} & \best{2.7} & \best{7.1} & 0.226 &
\best{0.785} & \best{16.5} & \best{30.1} & \best{0.106} & \best{0.878} \\
  \bottomrule
  \end{tabular}
  }
  \caption{Camera pose and depth accuracy with 10 input views, trained on RE10K dataset.}
  \label{tab:pose_re10k}
\end{table}

\begin{figure}[!h]
    \centering
    \includegraphics[width=\linewidth]{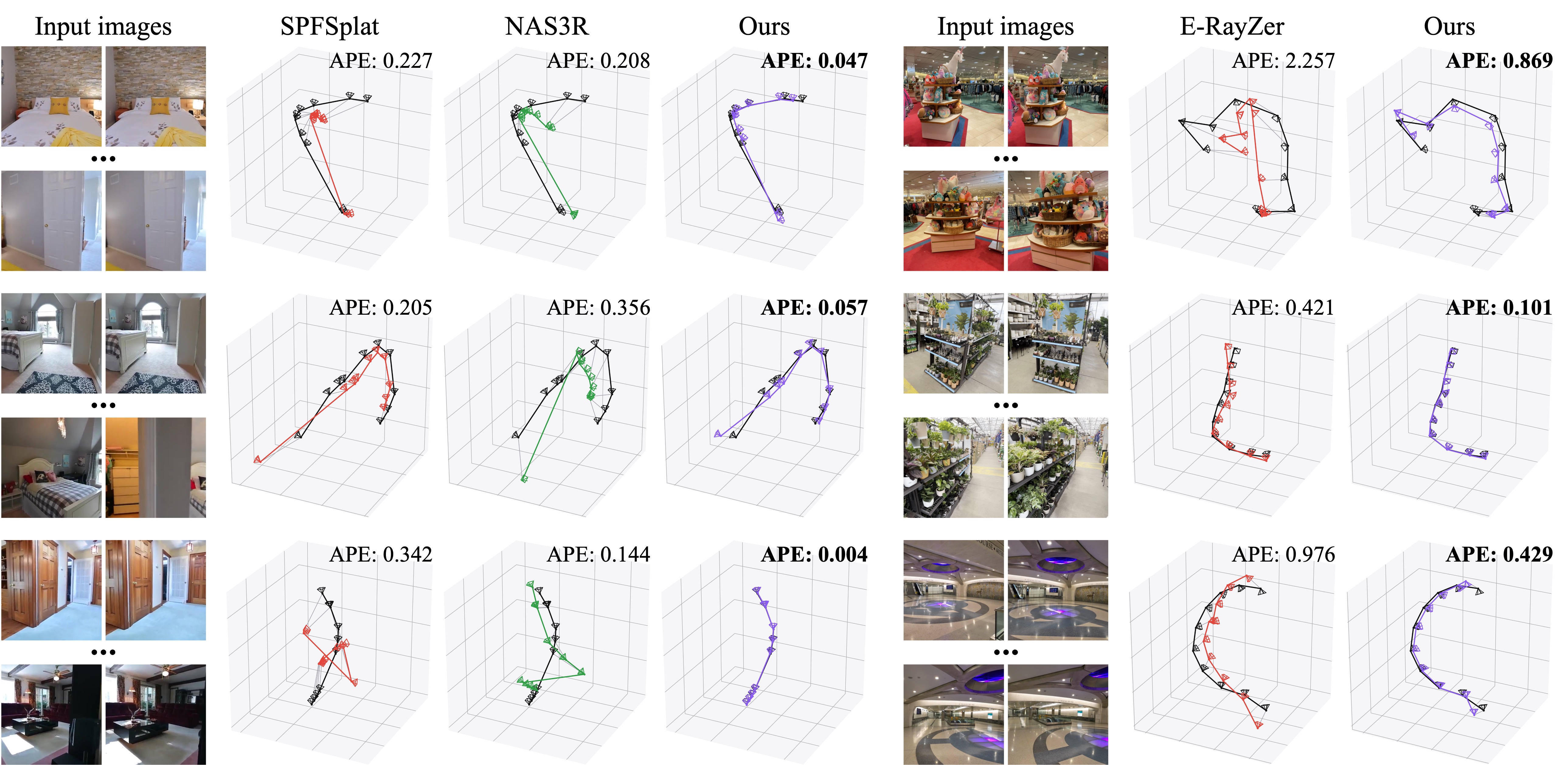}
    \vspace{-3mm}
    \caption{In-domain camera poses from 10 unposed views on RE10K (left) and DL3DV (right). Predictions are shown in color and ground truth in black, aligned by a similarity transform.}
    \label{fig:main_qual}
    \vspace{-2mm}
\end{figure}

\begin{figure}[!h]
    \centering
    \includegraphics[width=\linewidth]{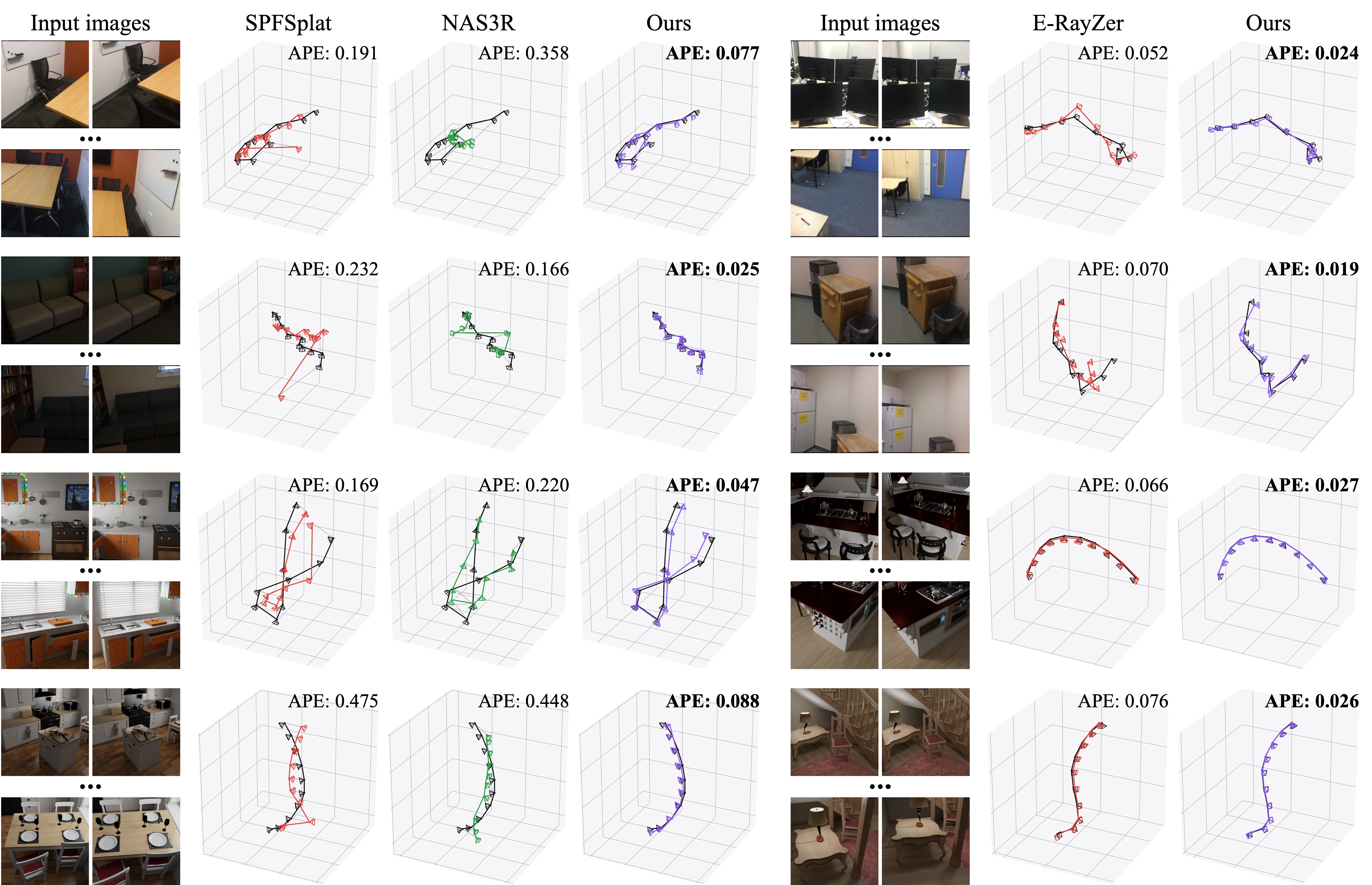}
    \vspace{-3mm}
    \caption{Out-of-domain camera poses on ScanNetv2 (top two rows) and NRGBD (bottom two rows), from models trained on RE10K (left) and DL3DV (right).}
    \label{fig:ood_qual}
    \vspace{-2mm}
\end{figure}

We also compare with E-RayZer~\citep{zhao2026erayzer}, with both models trained on DL3DV. Our model improves every reported pose and depth metric both in-domain and out-of-domain datasets (Tab.~\ref{tab:pose_dl3dv}). For depth, E-RayZer feeds only the context views to its geometry branch. Our backbone processes all views, but only the context views are decoded by the DPT head, so both methods are scored on the same context views. Fig.~\ref{fig:main_qual} and Fig.~\ref{fig:ood_qual} show the recovered camera trajectories on the in-domain and out-of-domain datasets, after Sim(3) alignment to the ground truth with the Umeyama method~\citep{umeyama1991least}.

\begin{table}[!h]
\vspace{-2mm}
\centering
\resizebox{\columnwidth}{!}{
\begin{tabular}{lccccccccccccc}
\toprule
\multirow{2}{*}{Method}  & \multirow{2}{*}{Params} & \multicolumn{2}{c}{DL3DV (trained)} &
\multicolumn{2}{c}{RE10K} & \multicolumn{4}{c}{ScanNetv2} & \multicolumn{4}{c}{NRGBD} \\
\cmidrule(lr){3-4} \cmidrule(lr){5-6} \cmidrule(lr){7-10} \cmidrule(lr){11-14}
& & RPA@3$\uparrow$ & @5$\uparrow$ & RPA@3$\uparrow$ & @5$\uparrow$ & RPA@3$\uparrow$ & @5$\uparrow$ & AbsRel$\downarrow$ & $\delta < 1.25\uparrow$ & RPA@3$\uparrow$ & @5$\uparrow$ & AbsRel$\downarrow$
& $\delta < 1.25\uparrow$ \\
\midrule
E-RayZer & 246M & 60.3 & 72.1 & 54.6 & 74.5 & 4.9 & 12.3 & 0.168 & 0.819 & 35.6 & 61.5 & 0.095 & 0.908 \\
Ours     & 183M & \best{70.0} & \best{79.7} & \best{57.7} & \best{75.3} & \best{6.6} & \best{16.7} & \best{0.125} &
\best{0.872} & \best{63.5} & \best{81.5} & \best{0.082} & \best{0.917} \\
\bottomrule
\end{tabular}
}
\caption{Camera pose and depth accuracy with 10 input views, trained on DL3DV dataset.}
\label{tab:pose_dl3dv}
\end{table}
\vspace{-2mm}

\begin{wraptable}{r}{0.5\textwidth}
  \vspace{-12pt}
  \centering
  \renewcommand{\arraystretch}{0.95}
  \resizebox{\linewidth}{!}{
  \begin{tabular}{llcccc}
  \toprule
  \multirow{2}{*}{Input} & \multirow{2}{*}{Method} & \multicolumn{2}{c}{ScanNetv2} & \multicolumn{2}{c}{NRGBD} \\
  \cmidrule(lr){3-4} \cmidrule(lr){5-6}
  & & AbsRel$\downarrow$ & $\delta < 1.25\uparrow$ & AbsRel$\downarrow$ & $\delta < 1.25\uparrow$ \\
  \midrule
  \multirow{2}{*}{Single image} & E-RayZer & 0.201 & 0.721 & 0.190 & 0.713 \\
                                & Ours     & \best{0.099} & \best{0.911} & \best{0.113} & \best{0.859} \\
  \midrule
  \multirow{2}{*}{Multi-view} & E-RayZer & 0.168 & 0.819 & 0.095 & 0.908 \\
                              & Ours     & \best{0.125} & \best{0.872} & \best{0.082} & \best{0.917} \\
  \midrule
  \multirow{2}{*}{Pose AUC@30$\uparrow$} & E-RayZer & \multicolumn{2}{c}{39.1} & \multicolumn{2}{c}{83.1} \\
                                         & Ours     & \multicolumn{2}{c}{\best{49.9}} & \multicolumn{2}{c}{\best{87.2}} \\
  \bottomrule
  \end{tabular}
  }
  \caption{Single- and multi-view depth performance. Both settings are evaluated on the same indices per scene after per-view median scaling.}
  \label{tab:single_view_depth}
  \vspace{-10pt}
\end{wraptable}

\paragraph{Single-image depth.}
We further evaluate depth when each frame is given to the model alone (Tab.~\ref{tab:single_view_depth}).
The network then receives only the reference camera token, so the pose head returns the identity pose together with a focal length predicted from the image, which we use to convert ray depth to $z$-depth. Here our model outperforms E-RayZer on both datasets. Our multi-view depth is more accurate than our single-image depth on NRGBD, where our pose estimation is accurate, but less accurate on ScanNetv2, where pose estimation is much harder. The same pattern holds across scenes within ScanNetv2. When we sort the scenes by per-scene pose error and split them into three groups, multi-view depth outperforms single-image depth in 45\% of the lowest-error scenes, 41\% of the middle group, and only 3\% of the highest-error scenes. Across all scenes, the multi-view AbsRel also increases with the pose error (Spearman $\rho = 0.45$). E-RayZer follows the same pattern, with 94\%, 88\%, and 56\% of scenes and $\rho = 0.61$. This suggests that pose estimation and multi-view depth depend on the same cross-view reasoning. 

\subsection{Analyses}
\label{analysis}
We conduct controlled analyses on the DL3DV dataset to isolate the effect of each design choice. Unless stated otherwise, we use the full-view attention design of Sec.~\ref{app:asym} and match model size and training compute across variants. We evaluate with $\{5,7,9,11\}$ views sampled from a fixed 48-frame span, reporting the average across view counts.

\textbf{Self-supervised pretraining.} We investigate whether self-supervised learning improves subsequent supervised training, using the same architecture and data throughout. The supervised model (SL) is trained from scratch for 200K iterations with dataset camera poses and a relative camera loss~\citep{wang2026pi}. The self-supervised model (SSL) is trained without geometric supervision for 100K iterations. The fine-tuned model (SSL+SFT) starts from SSL and receives the same supervision as SL for another 50K iterations, for 150K iterations in total.

As shown in Tab.~\ref{tab:ssl_sft}, SSL alone is comparable to SL, and is more accurate at strict pose thresholds despite using half the iterations. SSL+SFT surpasses both on every reported metric, while training for fewer total iterations than SL. Self-supervised learning thus provides an effective initialization for supervised geometry estimation.

\begin{table}[h]
    \centering
    \renewcommand{\arraystretch}{0.95}
    \resizebox{\columnwidth}{!}{
    \begin{tabular}{lcccccccccccc}
    \toprule
    \multirow{2}{*}{Method} & \multicolumn{4}{c}{DL3DV (trained)} & \multicolumn{4}{c}{ScanNetv2} & \multicolumn{4}{c}{NRGBD} \\
    \cmidrule(lr){2-5} \cmidrule(lr){6-9} \cmidrule(lr){10-13}
    & RPA@3$\uparrow$ & @5$\uparrow$ & @10$\uparrow$ & AUC@30$\uparrow$ & RPA@3$\uparrow$ & @5$\uparrow$ & @10$\uparrow$ & AUC@30$\uparrow$ & RPA@3$\uparrow$ & @5$\uparrow$ & @10$\uparrow$ & AUC@30$\uparrow$ \\
    \midrule
    SSL     & 66.6 & 81.7 & 91.7 & 88.0 & 1.2 & 5.2 & 17.6 & 29.8 & 16.5 & 37.8 & 74.3 & 72.3 \\
    SL      & 50.8 & 74.6 & 92.5 & 87.2 & 0.7 & 3.2 & 13.5 & 25.2 & 16.3 & 41.2 & 82.7 & 77.2 \\
    SSL+SFT & \best{73.4} & \best{89.3} & \best{97.2} & \best{92.4} & \best{3.4} & \best{10.1} & \best{28.4} & \best{41.4} & \best{33.3} & \best{62.0} & \best{87.9} & \best{81.9} \\
    \bottomrule
    \end{tabular}
    }
    \caption{Camera pose accuracy of self-supervised (SSL) and supervised (SL) training, and of self-supervised pretraining followed by supervised fine-tuning (SSL+SFT).}
    \label{tab:ssl_sft}
\end{table}

\textbf{Asymmetric feature alignment.} Eq.~\ref{eq:align} aligns the intermediate features of a sparse-context model to those of a dense-context model of the same scene. We compare against depth alignment, which instead applies a scale-invariant loss between the two depth maps. The alignment loss is introduced at 40\% of training and reaches full weight at 60\%. Tab.~\ref{tab:analy_align} shows that feature alignment improves every metric over training without alignment.

\begin{figure}[!h]
    \centering
    \includegraphics[width=\linewidth]{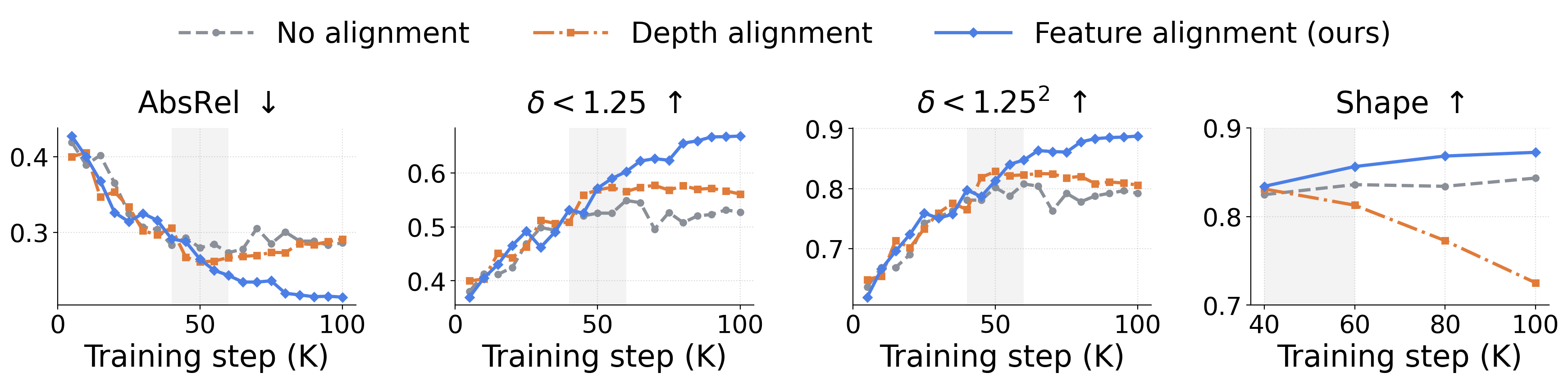}
    \vspace{-5mm}
    \caption{Single-view depth validation on DL3DV dataset over training steps. The shaded region marks the ramp-up of the alignment loss.}
    \label{fig:single_view_depth_align_analy}
    \vspace{-2mm}
\end{figure}

In contrast, depth alignment does not help. Fig.~\ref{fig:single_view_depth_align_analy} examines the metrics over training. Beyond standard metrics, we measure \emph{shape}, the Pearson correlation $\rho(\log\hat{z},\log z)$ against pseudo ground truth, which is invariant to the global scale and captures how depth is ordered across the image. Under depth alignment, shape steadily declines, so errors in the dense-context depth are passed on as wrong geometry, not just wrong scale. Feature alignment matches representations rather than depth values, leaving depth to the rendering loss.

\begin{table}[!h]
    \centering
    \small
    \setlength{\tabcolsep}{5pt}
    \renewcommand{\arraystretch}{0.95}
    \resizebox{\columnwidth}{!}{
    \begin{tabular}{lcccccccc}
    \toprule
    \multirow{2}{*}{Method} & \multicolumn{2}{c}{Camera pose} & \multicolumn{3}{c}{Multi-view depth} & \multicolumn{3}{c}{Single-view depth} \\
    \cmidrule(lr){2-3} \cmidrule(lr){4-6} \cmidrule(lr){7-9}
    & RPA@3 $\uparrow$ & RPA@5 $\uparrow$ & AbsRel $\downarrow$ & $\delta < 1.25 \uparrow$ & $\delta < 1.25^2\uparrow$ & AbsRel $\downarrow$ & $\delta < 1.25 \uparrow$ & $\delta < 1.25^2\uparrow$ \\
    \midrule
    No alignment & 66.6 & 81.7 & 0.195 & 0.758 & 0.901 & 0.230 & 0.638 & 0.850 \\
    Depth alignment & 63.4 & 81.7 & 0.244 & 0.692 & 0.854 & 0.248 & 0.651 & 0.844 \\
    Feature alignment (ours) & \best{69.8} & \best{84.6} & \best{0.191} & \best{0.768} & \best{0.905} & \best{0.191} & \best{0.729} & \best{0.901} \\
    \bottomrule
    \end{tabular}}
    \vspace{-2mm}
    \caption{Asymmetric context alignment analysis on DL3DV dataset.
    }
    \label{tab:analy_align}
    \vspace{-2mm}
\end{table}

\begin{figure}[!h]
    \centering
    \includegraphics[width=\linewidth]{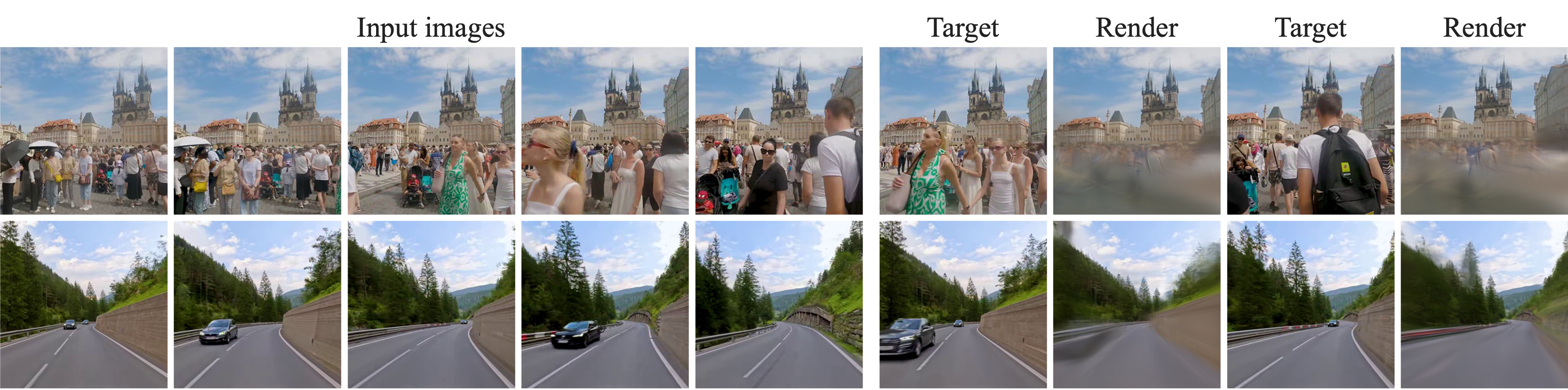}
    \vspace{-5mm}
    \caption{Moving people and vehicles are largely absent from the renderings at the target views, while the static background is reconstructed.}
    \label{fig:dynamic_example}
    \vspace{-2mm}
\end{figure}

\textbf{Dynamic monocular videos.} We also explore training on dynamic monocular videos from SpatialVID~\citep{wang2026spatialvid}, a large-scale dataset with diverse camera motion and moving content. We train the same model on DL3DV and SpatialVID with equal sampling per epoch, keeping the architecture, objective, and training schedule unchanged. The model thus still assumes a static scene, although many clips contain moving objects. Without any architectural change, the model tends to assign low opacity to Gaussians on moving objects (Fig.~\ref{fig:dynamic_example}). Since no static scene can explain these regions, their photometric error offers no consistent signal for camera pose, which likely leaves the static structure to drive camera estimation.

\begin{wraptable}{r}{0.5\textwidth}
  \vspace{-12pt}
  \centering
  \renewcommand{\arraystretch}{0.95}
  \resizebox{\linewidth}{!}{
  \begin{tabular}{lcccc}
  \toprule
  \multirow{2}{*}{Training data} & DL3DV & \multicolumn{3}{c}{SpatialVID} \\
  \cmidrule(lr){2-2} \cmidrule(lr){3-5}
  & benchmark & dyn\_low & dyn\_high & crowded \\
  \midrule
  DL3DV              & \textbf{80.7} / \textbf{87.1} & 36.9 / 70.0 & 25.4 / 55.7 & 24.1 / 55.1 \\
  DL3DV + SpatialVID & 79.1 / 85.8 & \textbf{74.7} / \textbf{86.0} & \textbf{73.9} / \textbf{84.0} & \textbf{87.4} / \textbf{89.2} \\
  \bottomrule
  \end{tabular}
  }
  \caption{Camera pose accuracy (RPA@5 / AUC@30, higher is better) of models trained on DL3DV only and on DL3DV with SpatialVID. SpatialVID test sets are grouped by the amount of dynamic content and use a 32-frame span. DL3DV uses a 48-frame span.}
  \label{tab:spatialvid}
  \vspace{-10pt}
\end{wraptable}

\begin{figure}[!h]
    \centering
    \includegraphics[width=\linewidth]{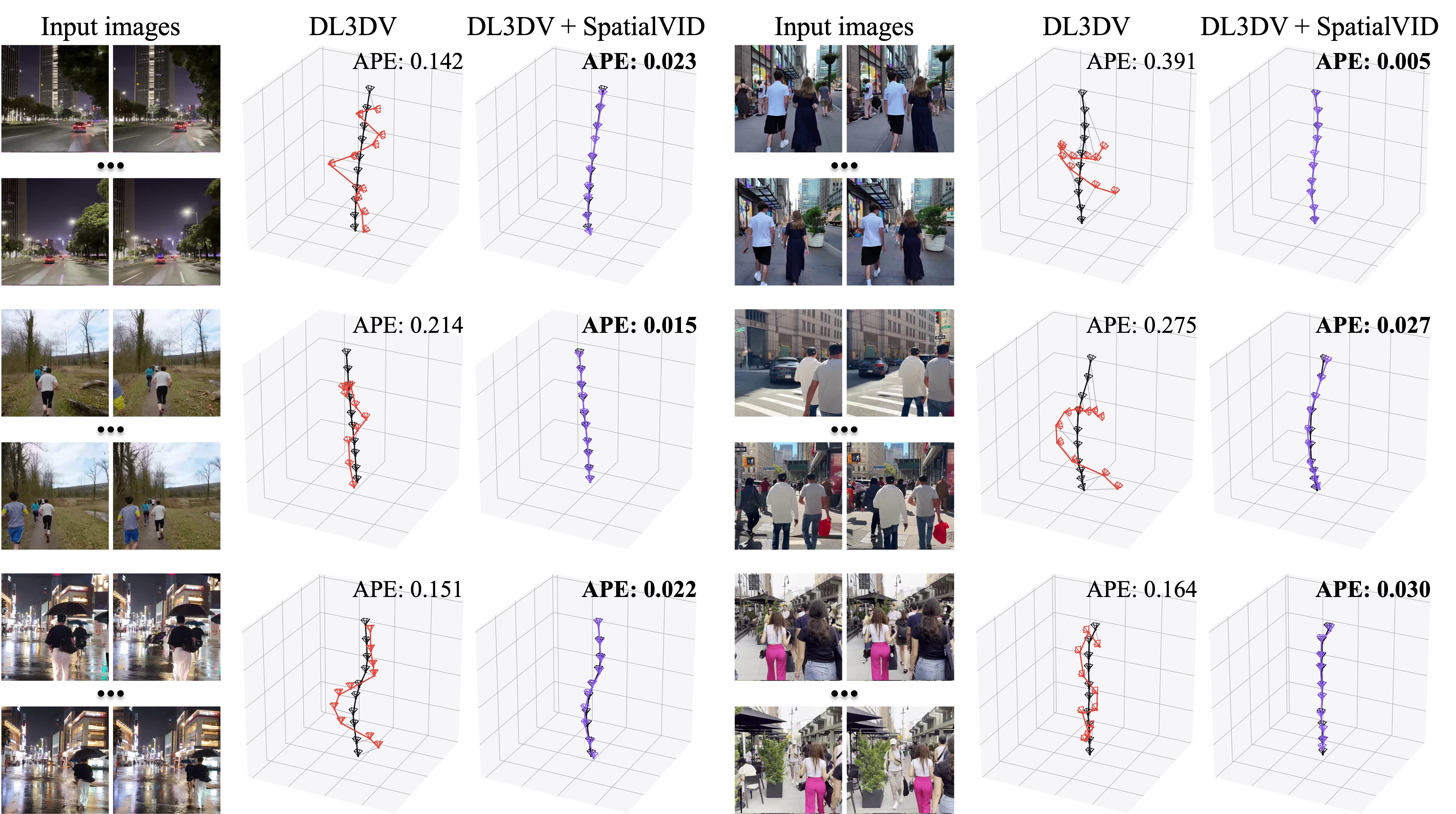}
    \vspace{-3mm}
    \caption{Camera poses on held-out SpatialVID videos with many moving objects (left) and crowds (right), from models trained on DL3DV only and on DL3DV with SpatialVID.}
    \label{fig:main_dynamic}
    \vspace{-2mm}
\end{figure}

To examine how dynamic content affects pose estimation, we construct three disjoint test sets of 100 held-out SpatialVID scenes each from the dataset's metadata: near-static scenes (\textit{dyn\_low}) and highly dynamic scenes (\textit{dyn\_high}), taken from the lower and upper deciles of the dynamic ratio, and crowded scenes (\textit{crowded}). For each scene, we sample 10 views over a 32-frame span and compare against the same model trained on DL3DV alone (Tab.~\ref{tab:spatialvid}). Training with SpatialVID improves pose accuracy in all three groups, including near-static scenes, suggesting that the gain may stem partly from greater camera motion diversity rather than solely from robustness to dynamic content. On DL3DV, the two models perform comparably. Fig.~\ref{fig:main_dynamic} shows the recovered trajectories.

\section{Discussion}
\textbf{Design motivation.} Prior pose-free methods restrict how held-out views enter the backbone, either by excluding them from geometry prediction or by preventing context views from attending to them. This choice is natural when rendering quality is the goal, since observing a held-out view could reveal what the model is asked to render. Our goal is geometry, with view synthesis serving only as the training signal. We keep the context and held-out split, but apply it to the 3D scene, so a held-out view is never rendered from its own prediction while still contributing correspondence in the shared backbone. For this reason, rendering metrics are not our primary basis of comparison.

All designs receive the same images during training and differ only in how they use them. In restricted designs, held-out views cannot inform the context-view predictions from which the scene is built. In ours, every view informs every other. The gain therefore comes not from more data but from using the same views more effectively.

\textbf{Conclusion and future work.} We study self-supervised geometry learning with rendering as the training objective and geometric accuracy as the outcome. Our central finding is that the asymmetry needed for self-supervision belongs in the 3D representation rather than in the network. Building on this, aligning sparse-context features with dense-context features turns view density into a further learning signal, and aligning with a frozen vision foundation model adds a semantic prior that pixels alone do not provide. Together, these choices form a simple design that improves camera pose and depth without geometric annotations.

A key opportunity is to learn from visual data beyond annotation pipelines. Much of this data is casually captured monocular video, where the scene itself changes between frames, while our framework assumes a static scene. Our preliminary experiment suggests that the static model already tolerates some motion by assigning low opacity to moving content, which improves camera pose estimation on dynamic videos. This does not model motion, however, and moving objects are simply left out of the reconstruction. A natural next step can be an explicit dynamic representation that separates object motion from camera motion, either with Gaussians that move over time~\citep{wu20244dgs, xu20254dgt} or with a separate reconstruction at each timestep in a bullet-time formulation~\citep{liang2024bullet}. Such a representation offers a path toward self-supervised geometry learning from the far larger supply of monocular video.

\bibliography{iclr2027_conference}
\bibliographystyle{iclr2027_conference}

\clearpage
\appendix
\section{Appendix}

\subsection{Additional Details}
\label{sec:addition_experiments}
\textbf{Training details.} We train the main model for 100K iterations on RE10K~\citep{zhou2018stereo} and 140K iterations on DL3DV~\citep{ling2024dl3dv} on 8 NVIDIA B200 GPUs. The analysis variants are trained for 100K iterations on DL3DV on 2 NVIDIA RTX A6000 GPUs. Images are resized to $256\times256$ and tokenized with a patch size of 16, giving 256 tokens per view. Each batch uses a single view count from $\{5,7,9,11\}$, split into $(N{+}1)/2$ context and $(N{-}1)/2$ target views, with per-device batch sizes of $\{24,20,16,12\}$ to keep the number of tokens per step roughly constant. Following RayZer~\citep{jiang2025rayzer}, we widen the frame span from which views are sampled over the first 80K iterations, moving from shorter to longer baselines, and keep it fixed afterward. The span grows from 24--96 to 48--192 frames on RE10K, from 24--48 to 32--54 on DL3DV, and from 24--32 to 32--48 for the analysis variants. We use AdamW~\citep{loshchilov2017decoupled} with $\beta_1=0.9$, $\beta_2=0.95$, and weight decay $0.05$, together with a cosine schedule and 3000 warm-up steps. The learning rate is $2\times10^{-4}$ for the main model and $1\times10^{-4}$ for the analysis variants. Training uses bf16 and FlashAttention-2~\citep{dao2024flashattention}, with gradients clipped to a norm of $1.0$.

\textbf{Evaluation details.} We evaluate on 5,601 RE10K scenes~\citep{zhou2018stereo}, 140 DL3DV scenes~\citep{ling2024dl3dv}, 9 NRGBD scenes~\citep{azinovic2022nrgbd}, and 97 ScanNetv2 scenes~\citep{dai2017scannetv2}. We evaluate SPFSplat and NAS3R with their official evaluation code, which applies
no masking at test time. In this comparison, depth is predicted and scored on all input views for both the baselines and our model. For E-RayZer, poses are estimated from all images and depth from the context views only. Accordingly, when comparing with E-RayZer, our DPT head decodes only the context views, and depth is scored on the same context views for both methods.

\begin{figure}[!h]
    \centering
    \vspace{-2mm}     
    \includegraphics[width=0.9\linewidth]{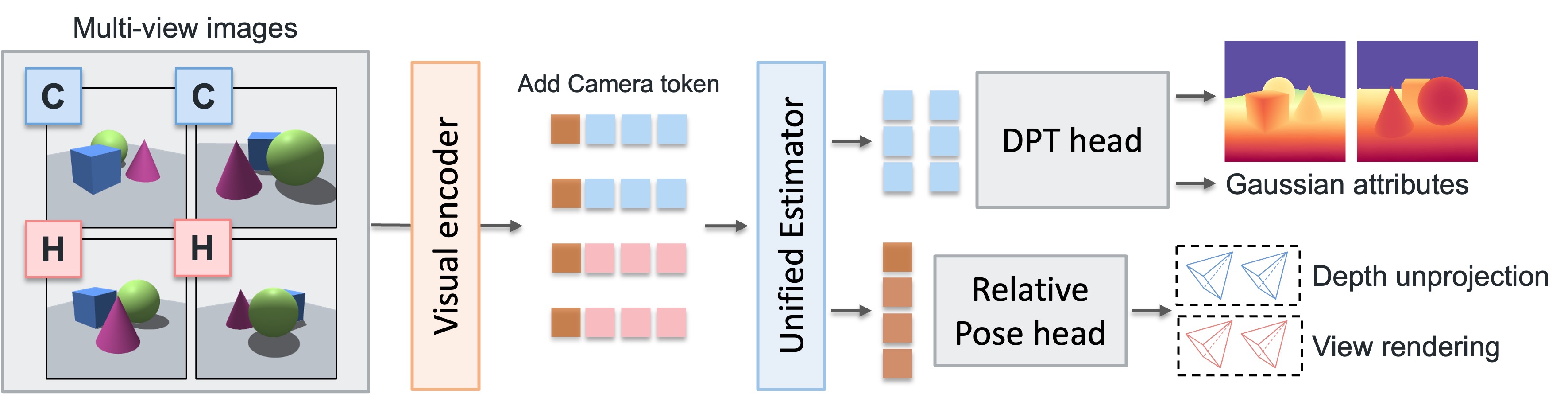}
    \caption{Overall architecture, with a shared backbone for camera and geometry prediction.}
    \label{fig:arch}
    \vspace{-2mm}       
\end{figure}

\textbf{Architecture.} Fig.~\ref{fig:arch} illustrates the pipeline. A visual encoder tokenizes each view, and a learnable camera token is prepended to its patch tokens. Tokens from both context (C) and held-out (H) views enter a unified estimator with alternating view-wise and global attention, so every view can interact with every other. The resulting patch tokens feed a DPT head~\citep{ranftl2021dpt}, which reads four intermediate estimator layers to predict per-pixel depth and the remaining Gaussian attributes. The camera tokens feed a relative pose head, which regresses each pose relative to the middle input view. Context-view depths are unprojected with the predicted cameras to form the Gaussian scene, which is then rendered at the predicted held-out poses. Since the asymmetry is imposed in the 3D representation, attention in the estimator remains unrestricted.

\subsection{Additional Analyses}

\begin{wrapfigure}{r}{0.58\textwidth}
  \begin{center}
  \vspace{-12mm}   
    \includegraphics[width=0.92\linewidth, trim=0 0 0 10]{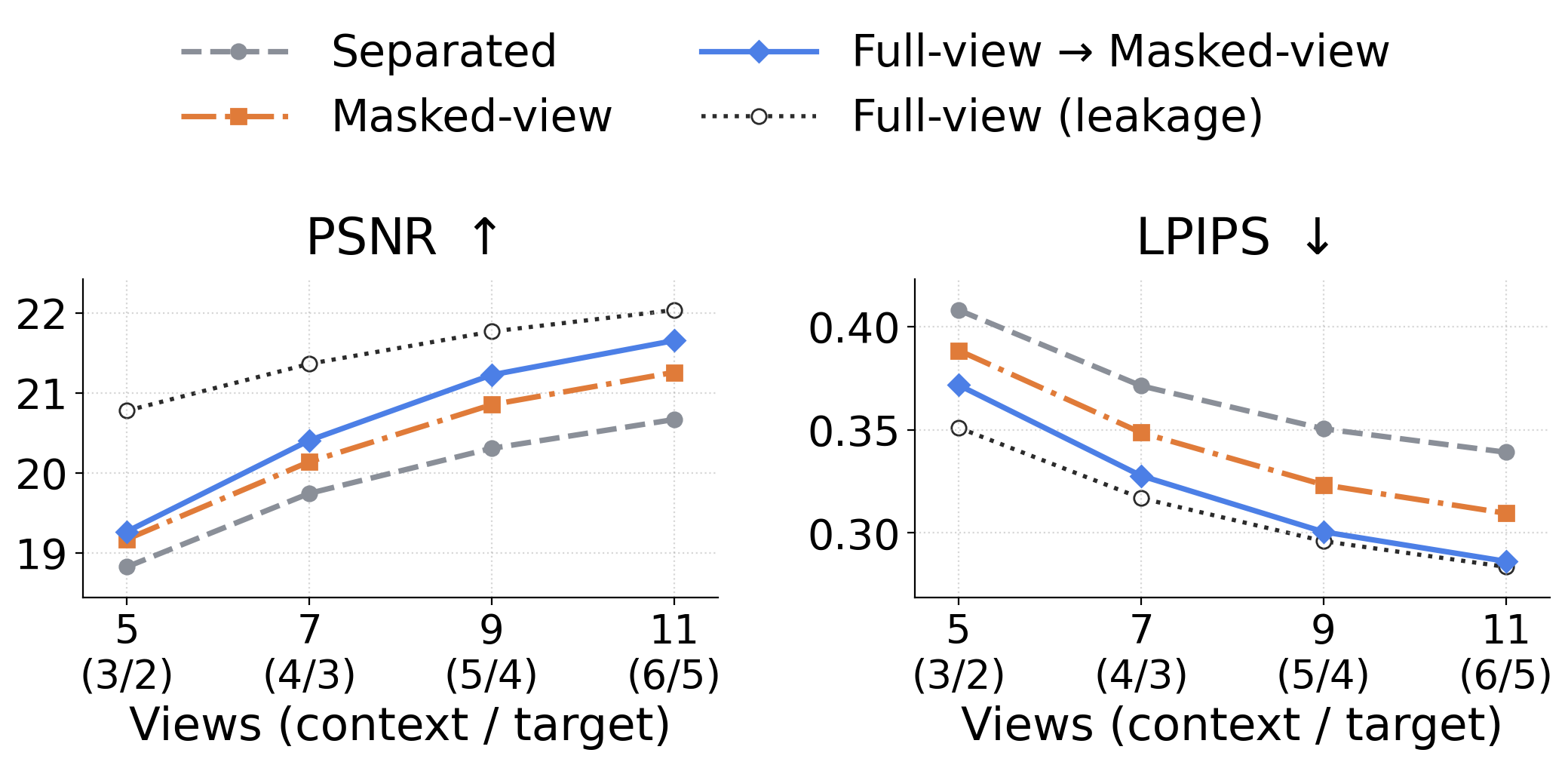}
    \vspace{-2mm}   
    \caption{View synthesis evaluation. Rendering metrics on DL3DV dataset as the number of input views grows, with the context and target split shown in parentheses.}
    \label{fig:input_asym_psnr}
    \end{center}
    \vspace{-8mm}   
\end{wrapfigure}

\textbf{View synthesis under full-view attention.} Full-view attention lets target images inform context tokens at test time, so its rendering metrics are not directly comparable to those of restricted designs. To evaluate view synthesis without this leakage, we train with full-view attention for the first 80\% of iterations and switch to masked attention for the remainder, then evaluate with masked attention. We compare against a model trained with masked attention throughout under the same total schedule. The switched model renders consistently better (Fig.~\ref{fig:input_asym_psnr}), suggesting that the geometry learned under full-view attention carries over once leakage is removed.

\textbf{Effect of context in the posed setting.} The main experiments predict cameras jointly with geometry, so the benefit of full-view attention could come from better pose estimation rather than better geometry. To separate the two, we repeat the comparison with ground-truth cameras on DL3DV dataset. Both variants build Gaussians from context views only and differ in a single choice. FULL feeds all views, context and target, into the encoder during training, while MASK feeds only the context views. If target views offered a shortcut, the geometry predicted by FULL would become biased toward the targets and degrade. If they instead supply correspondence, FULL should recover better geometry. At test time, both models encode only the context views, so their inputs are identical. As a diagnostic, we also evaluate FULL with all views encoded, as during training, to see how much it relies on target views. Depth is always scored on the same context views.

\begin{figure}[!h]
    \vspace{-3mm}
    \centering
    \includegraphics[width=0.8\linewidth]{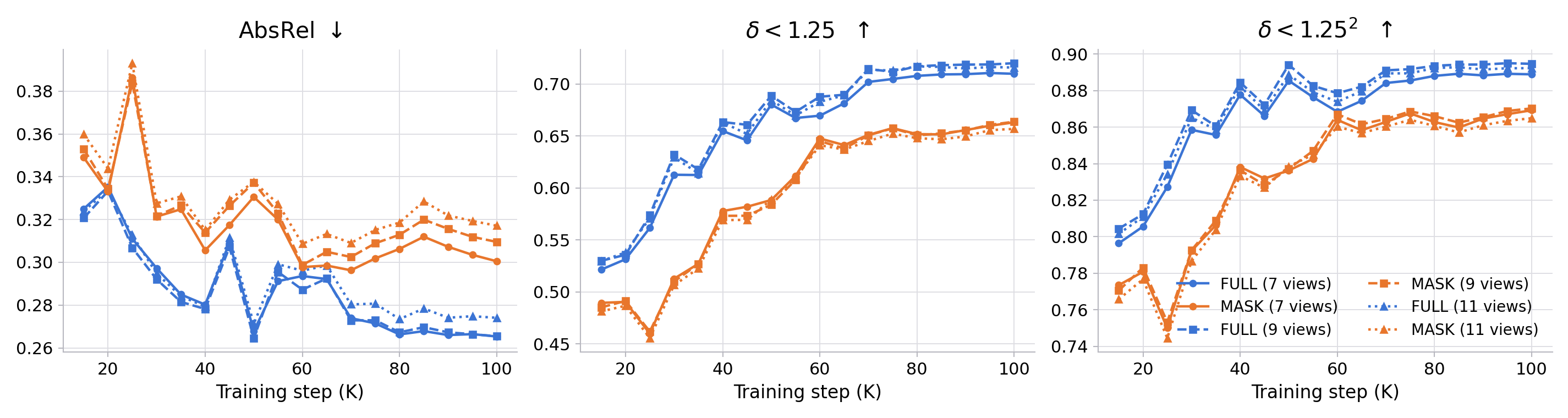}
    \caption{Depth validation over training in the posed setting.}
    \label{fig:posed_graph}
    \vspace{-3mm}
\end{figure}

\begin{table}[!h]
    \centering
    \resizebox{\textwidth}{!}{
    \begin{tabular}{lcccccccccc}
    \toprule
    \multirow{2}{*}{Views} & \multicolumn{5}{c}{FULL} & \multicolumn{5}{c}{MASK} \\
    \cmidrule(lr){2-6} \cmidrule(lr){7-11}
    & AbsRel $\downarrow$ & $\delta < 1.25 \uparrow$ & $\delta < 1.25^2\uparrow$ & PSNR $\uparrow$ & LPIPS $\downarrow$ & AbsRel $\downarrow$ & $\delta < 1.25 \uparrow$ & $\delta < 1.25^2\uparrow$ & PSNR $\uparrow$ & LPIPS $\downarrow$ \\
    \midrule
    7 (4/3)  & \best{0.271} {\color{gray}(0.274)} & \best{0.729} {\color{gray}(0.743)} & \best{0.887} {\color{gray}(0.892)} & 20.42 {\color{gray}(21.32)} & 0.365 {\color{gray}(0.328)} & 0.306 & 0.693 & 0.873 & \best{21.17} & \best{0.328} \\
    9 (5/4)  & \best{0.276} {\color{gray}(0.281)} & \best{0.734} {\color{gray}(0.746)} & \best{0.887} {\color{gray}(0.892)} & 20.92 {\color{gray}(21.74)} & 0.344 {\color{gray}(0.312)} & 0.319 & 0.689 & 0.871 & \best{21.68} & \best{0.311} \\
    11 (6/5) & \best{0.276} {\color{gray}(0.286)} & \best{0.743} {\color{gray}(0.748)} & \best{0.892} {\color{gray}(0.894)} & 21.27 {\color{gray}(21.98)} & 0.329 {\color{gray}(0.301)} & 0.319 & 0.688 & 0.872 & \best{22.00} & \best{0.299} \\
    \bottomrule
    \end{tabular}
    }
    \caption{Effect of full interaction in the posed setting. Views are listed as total (context/target), and depth is scored on the context views. FULL encodes only the context views at test time, matching MASK. Gray values in parentheses are from FULL with all views encoded, shown for reference.}
    \label{tab:posed}
\end{table}

\emph{More context improves geometry.} With identical inputs at test time, FULL recovers better depth than MASK on every metric and at every view count (Tab.~\ref{tab:posed}), and the gap holds throughout training (Fig.~\ref{fig:posed_graph}). Since the two models differ only in whether target views entered the encoder during training, the gain comes from training with unrestricted interaction.

\textbf{Semantic guidance.} Tab.~\ref{tab:semantic_guidance} isolates the effect of aligning encoder features with a frozen vision foundation model. We use either Pixio~\citep{yang2026pixio}, a masked autoencoder, or DINOv3~\citep{simeoni2025dinov3}, each serving both to initialize the encoder and as the frozen model. Pretrained initialization already matters substantially, as training from random initialization lowers every metric by a wide margin. Semantic guidance improves on this for both encoders, even though the encoder starts from the same pretrained weights it is aligned to. The largest gain is in camera pose, particularly at strict thresholds, and depth improves as well. This suggests that initialization alone does not retain the pretrained prior under the rendering loss, and that keeping it as a training signal helps preserve it. Adding a projection before the alignment performs similarly, so we use the version without it. Pixio yields clearly better pose than DINOv3 at comparable depth, and we use it in all other experiments.

\begin{table}[!h]
    \centering
    \renewcommand{\arraystretch}{0.95}
    \resizebox{1.0\columnwidth}{!}{
        \begin{tabular}{llccccccc}
        \toprule
        \multirow{2}{*}{Backbone} & \multirow{2}{*}{Method} & \multicolumn{3}{c}{Camera pose} & \multicolumn{2}{c}{Depth} & \multicolumn{2}{c}{NVS} \\
        \cmidrule(lr){3-5} \cmidrule(lr){6-7} \cmidrule(lr){8-9}
        & & RPA@3 $\uparrow$ & RPA@5 $\uparrow$ & RPA@10 $\uparrow$ & AbsRel $\downarrow$ & $\delta < 1.25 \uparrow$ & PSNR $\uparrow$ & LPIPS $\downarrow$ \\
        \midrule
        \color{gray}-- & \color{gray}w/o Pretrained init & \color{gray}36.2 & \color{gray}56.6 & \color{gray}77.3 & \color{gray}0.247 & \color{gray}0.652 & \color{gray}19.82 & \color{gray}0.438 \\
        \midrule
        \multirow{3}{*}{Pixio (base)} & Baseline & 66.6 & 81.7 & 91.7 & 0.195 & 0.758 & 21.48 & 0.312 \\
         & w/ Sem. guidance & \best{82.3} & \best{90.8} & 96.0 & 0.185 & \best{0.790} & \best{21.55} & \best{0.305} \\
         & w/ Sem. guidance + Proj. & 81.6 & 90.7 & \best{96.2} & 0.189 & 0.789 & 21.52 & \best{0.305} \\
        \midrule
        \multirow{3}{*}{DINOv3 (base)} & Baseline & 46.8 & 64.3 & 80.0 & 0.217 & 0.718 & 20.23 & 0.361 \\
         & w/ Sem. guidance & 62.4 & 79.3 & 91.2 & \best{0.181} & 0.778 & 20.40 & 0.361 \\
         & w/ Sem. guidance + Proj. & 60.8 & 78.1 & 90.3 & 0.183 & 0.773 & 20.26 & 0.367 \\
        \bottomrule
        \end{tabular}
    }
    \vspace{-2mm}
    \caption{Ablation of semantic guidance on DL3DV dataset. NVS metrics are
    reported as a reference only, since target views are processed together with
    the context under full-view attention.}
    \label{tab:semantic_guidance}
    \vspace{-2mm}
\end{table}

\textbf{3D probing.} We test whether semantic guidance improves the learned representation itself. Each self-supervised model (Pixio variants) from Tab.~\ref{tab:semantic_guidance} is frozen, and only its prediction heads are trained with 3D supervision. We train the heads on BlendedMVS~\citep{yao2020blendedmvs} and evaluate on 43 held-out scenes, and train them on ScanNet++~\citep{yeshwanth2023scannet++} and evaluate on 97 ScanNetv2~\citep{dai2017scannetv2} scenes. Without pretrained initialization, performance is lower on every metric. Semantic guidance improves on the self-supervised baseline on both datasets, in depth as well as pose, with the largest gains in camera pose (Tab.~\ref{tab:probe}). Since only the heads are trained, these gains reflect a better representation rather than a better fit to the probing data.

\begin{table}[!h]
    \centering
    \renewcommand{\arraystretch}{0.90}
    \resizebox{\columnwidth}{!}{
    \begin{tabular}{lcccccccccc}
    \toprule
    \multirow{2}{*}{Method} & \multicolumn{5}{c}{BlendedMVS} & \multicolumn{5}{c}{ScanNetv2} \\
    \cmidrule(lr){2-6} \cmidrule(lr){7-11}
    & AbsRel $\downarrow$ & $\delta < 1.25 \uparrow$ & RPA@5 $\uparrow$ & RPA@15 $\uparrow$ & RPA@30 $\uparrow$
    & AbsRel $\downarrow$ & $\delta < 1.25 \uparrow$ & RPA@5 $\uparrow$ & RPA@15 $\uparrow$ & RPA@30 $\uparrow$ \\
    \midrule
    \color{gray}w/o Pretrained init & \color{gray}0.162 & \color{gray}0.800 & \color{gray}4.2 & \color{gray}16.9 & \color{gray}26.7 & \color{gray}0.179 & \color{gray}0.745 & \color{gray}0.7 & \color{gray}9.8 & \color{gray}28.7 \\
    \midrule
    Baseline                  & 0.113 & 0.891 & 9.1 & 25.0 & 37.7 & 0.120 & 0.873 & 2.9 & 21.0 & 47.7 \\
    w/ Sem.\ guidance         & \best{0.094} & \best{0.919} & \best{12.6} & \best{33.0} & \best{44.3} & \best{0.078} & \best{0.943} & 6.2 & \best{39.5} & \best{67.6} \\
    w/ Sem.\ guidance + Proj. & 0.096 & \best{0.919} & 11.8 & 28.5 & 41.5 & 0.082 & 0.932 & \best{6.4} & 35.7 & 63.0 \\
    \bottomrule
    \end{tabular}
    }
    \vspace{-2mm}
    \caption{3D probing with semantic guidance. Self-supervised models from Tab.~\ref{tab:semantic_guidance} are frozen, and only the output heads are trained with 3D supervision on each dataset. Each scene uses 10 frames spanning 24 frames on BlendedMVS and 64 frames on ScanNetv2.}    
    \label{tab:probe}
    \vspace{-2mm}
\end{table}

\begin{wraptable}{r}{0.48\textwidth}
  \centering
  \vspace{-3mm}
  \renewcommand{\arraystretch}{0.90}
  \resizebox{\linewidth}{!}{
    \begin{tabular}{lcccc}
    \toprule
    Method & EPE $\downarrow$ & $<$1px $\uparrow$ & $<$3px $\uparrow$ & $<$5px $\uparrow$ \\
    \midrule
    \color{gray}w/o Pretrained init & \color{gray}4.05 & \color{gray}25.0 & \color{gray}68.0 & \color{gray}81.9 \\
    \midrule
    Baseline & 3.16 & 32.0 & 74.7 & 86.7 \\
    w/ Sem. guidance & 2.74 & 31.4 & 77.4 & 89.6 \\
    w/ Sem. guidance + Proj. & \best{2.66} & \best{32.3} & \best{78.7} & \best{90.2} \\
    \bottomrule
    \end{tabular}
  }
  \vspace{-2mm}
  \caption{Flow probing on StaticThings3D with a frozen backbone, on non-occluded pixels.}
  \label{tab:flow_probe}
  \vspace{-4mm}
\end{wraptable}

\textbf{Flow estimation.} We also probe dense two-view correspondence. Following UFM~\citep{zhang2025ufm}, we train only a DPT flow head on top of each frozen model on StaticThings3D~\citep{schroppel2022benchmark}, and evaluate on non-occluded pixels of 200 image pairs from 50 held-out sequences. Without pretrained initialization, performance is again lower on every metric and semantic guidance improves on the baseline (Tab.~\ref{tab:flow_probe}).

\begin{figure}[!h]
    \centering
    \includegraphics[width=\linewidth]{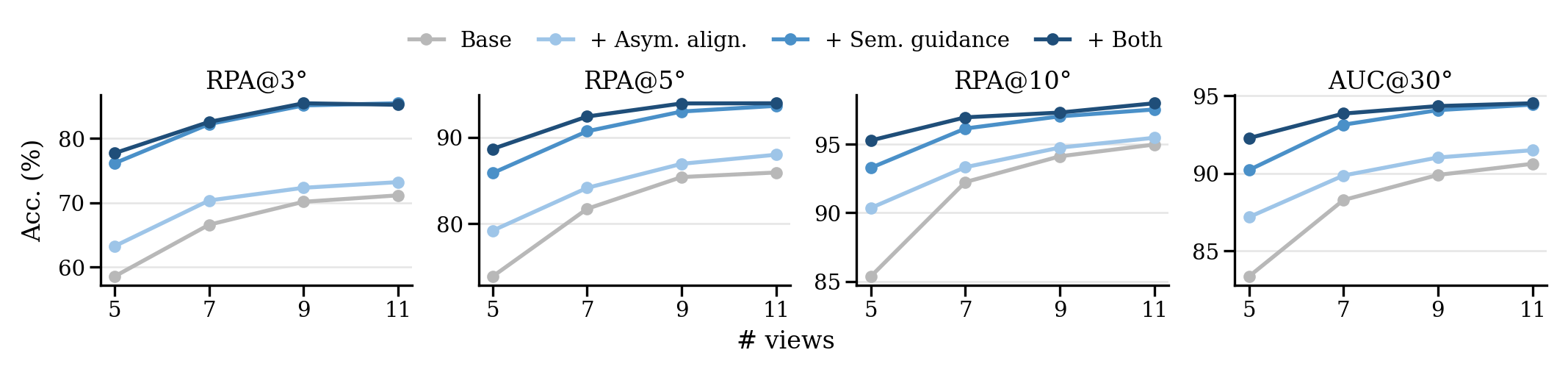}
    \vspace{-6mm}
    \caption{Camera pose accuracy on DL3DV when adding semantic guidance and asymmetric context alignment to the self-supervised baseline, averaged over 5, 7, 9, and 11 input views.}
    
    \label{fig:pose_abl_pose}
    \vspace{-2mm}
\end{figure}

\textbf{Combining the two signals.} Tab.~\ref{tab:ablation_stack} adds asymmetric context alignment and semantic guidance to the self-supervised baseline, separately and together. Each signal improves every metric over the baseline, but to different degrees. Semantic guidance gives the larger gain on most metrics, especially camera pose at strict thresholds, whereas the gains of asymmetric feature alignment concentrate in single-view depth, consistent with its role of transferring dense-context features to sparse inputs. The two are complementary. Enabling both yields the more accurate camera pose and the best single-view depth. Multi-view depth remains on par with semantic guidance alone, so asymmetric feature alignment adds its benefit in pose and in depth from sparse context rather than in dense multi-view setting. Fig.~\ref{fig:pose_abl_pose} shows per-view pose evaulation results.

\begin{table}[!h]
    \centering
    \renewcommand{\arraystretch}{0.90}
    \resizebox{1.0\columnwidth}{!}{
        \begin{tabular}{cccccccccc}
        \toprule
        \multirow{2}{*}{Sem. guid.} & \multirow{2}{*}{Asym. align.} & \multicolumn{4}{c}{Camera pose} & \multicolumn{2}{c}{Multi-view depth} & \multicolumn{2}{c}{Single-view depth} \\
        \cmidrule(lr){3-6} \cmidrule(lr){7-8} \cmidrule(lr){9-10}
        & & RPA@3 $\uparrow$ & RPA@5 $\uparrow$ & RPA@10 $\uparrow$ & AUC@30 $\uparrow$ & AbsRel $\downarrow$ & $\delta < 1.25$ $\uparrow$ & AbsRel $\downarrow$ & $\delta < 1.25$ $\uparrow$ \\
        \midrule
         &  & 66.6 & 81.7 & 91.7 & 88.0 & 0.195 & 0.758 & 0.230 & 0.638 \\
         & \checkmark & 69.8 & 84.6 & 93.5 & 89.9 & 0.191 & 0.768 & 0.191 & 0.729 \\
        \checkmark &  & 82.3 & 90.8 & 96.0 & 93.0 & \best{0.185} & \best{0.790} & 0.173 & 0.757 \\
        \checkmark & \checkmark & \best{82.8} & \best{92.2} & \best{96.9} & \best{93.7} & 0.188 & \best{0.790} & \best{0.165} & \best{0.799} \\
        \bottomrule
        \end{tabular}
    }
    \vspace{-2mm}
    \caption{Stacking semantic guidance and asymmetric context alignment on DL3DV.}
    \label{tab:ablation_stack}
    \vspace{-4mm}
\end{table}

\textbf{Comparison with neural rendering methods.} We compare our explicit 3D representation with methods that render target views through a learned decoder~\citep{sajjadi2022srt, jin2024lvsm} (Tab.~\ref{tab:neural_rendering}). These methods train with either interpolative targets, sampled between context views~\citep{jiang2025rayzer, wang2025less}, or predictive targets that may lie beyond them~\citep{mitchel2026xfactor, prestel2026rayder}. To control for differences in data and training, we implement one variant of each within our framework. Both use the same 12-layer backbone as our model followed by a 12-layer transformer decoder, and receive the same training data and compute. For the predictive variant, we follow the random-order autoregressive training of RayDer~\citep{prestel2026rayder}. Both variants achieve higher rendering scores, but neither predicts camera poses that correspond to physical camera motion, and neither produces explicit depth. XFactor~\citep{mitchel2026xfactor} evaluates whether learned latent poses transfer across scenes, which tests their consistency rather than their physical accuracy. Since our goal is geometry, we evaluate against the ground truth.

\begin{table}[!h]
    \centering
    \renewcommand{\arraystretch}{0.9}
    \resizebox{1.0\columnwidth}{!}{
        \begin{tabular}{lcccccccc}
        \toprule
        \multirow{2}{*}{Method} & \multicolumn{2}{c}{NVS} & \multicolumn{4}{c}{Camera pose} & \multicolumn{2}{c}{Depth} \\
        \cmidrule(lr){2-3} \cmidrule(lr){4-7} \cmidrule(lr){8-9}
        & PSNR $\uparrow$ & LPIPS $\downarrow$ & RPA@3 $\uparrow$ & RPA@5 $\uparrow$ & RPA@10 $\uparrow$ & AUC@30 $\uparrow$ & AbsRel $\downarrow$ & $\delta < 1.25 \uparrow$ \\
        \midrule
        Neural rendering (interpolative) & \best{22.28} & \best{0.258} & 0.0 & 0.0 & 1.1 & 0.26 & -- & -- \\
        Neural rendering (predictive) & 22.10 & \best{0.258} & 0.0 & 0.3 & 0.8 & 0.21 & -- & -- \\
        Ours & \color{gray}21.75  & \color{gray}0.280 & \best{85.7} & \best{94.1} & \best{98.5} & \best{95.12} & \best{0.230} & \best{0.775} \\
        \bottomrule
        \end{tabular}
    }
    \vspace{-2mm}
    \caption{Comparison with neural rendering methods on DL3DV dataset (8 views spanning 32 frames; 4 context + 4 target). The two neural rendering variants differ in whether target views are sampled between or beyond the context views during training. Our NVS metrics are grayed out, as target views are processed together with the context under full-view attention.}
    \label{tab:neural_rendering}
    \vspace{-2mm}
\end{table}

\textbf{More evaluation on dynamic videos.} We also evaluate the two models from the dynamic-video experiment, one trained on DL3DV only and one on DL3DV with SpatialVID, on ORBIT~\citep{sabour2026orbit}. Neither model is trained on this benchmark. ORBIT renders perspective views from 360$^\circ$ videos along virtual camera trajectories and provides ground-truth camera poses. Many clips also contain moving people and objects. We used 44 clips, which come from 21 video segments of 16 YouTube videos and have 203 to 1{,}024 frames each. We split each clip into consecutive, non-overlapping windows of $S$ frames, keep only windows that fit entirely within the clip, and evaluate each window as a separate sample. From each window we sample 10 views evenly and pass all of them to the model. This gives 330 windows for $S{=}64$ and 220 for $S{=}96$.

\begin{figure}[!h]
    \centering
    \includegraphics[width=\linewidth]{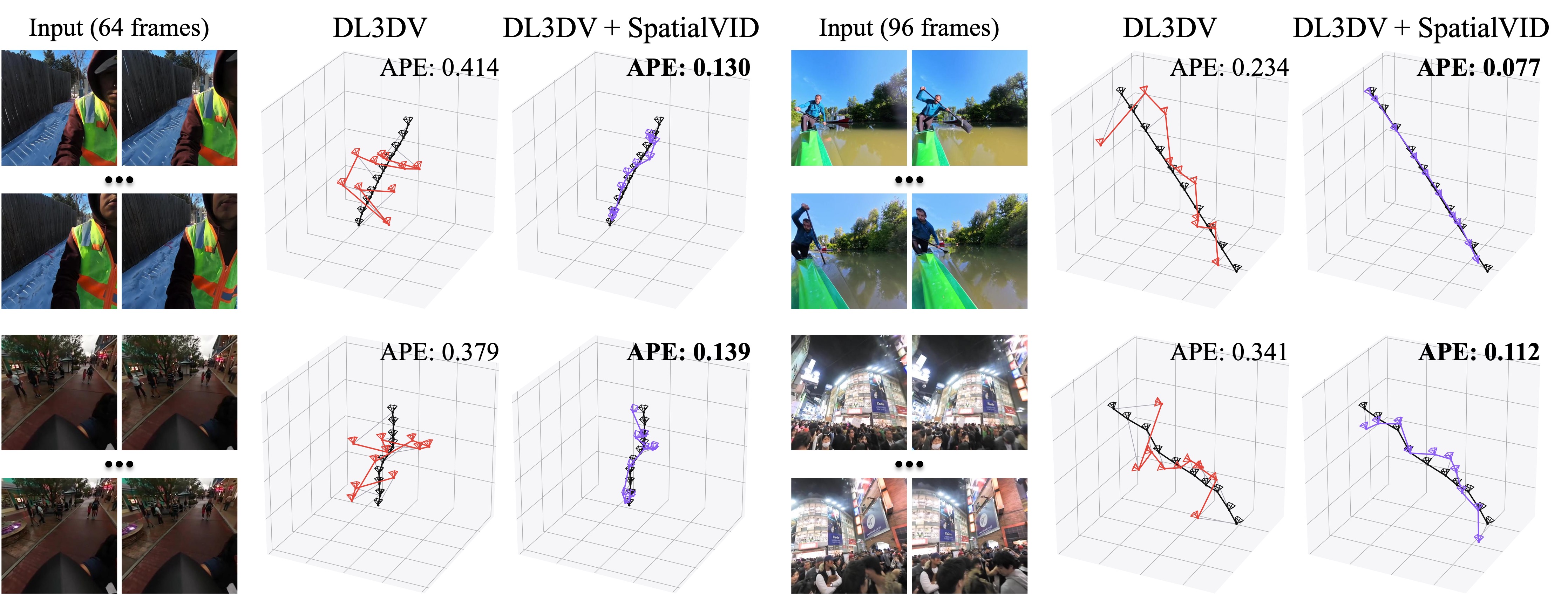}
    \vspace{-3mm}
    \caption{Camera poses on ORBIT windows spanning 64 frames (left) and 96 frames (right), from models trained on DL3DV only and on DL3DV with SpatialVID. Predictions are shown in color and ground truth in black, aligned by a similarity transform.
    }
    \label{fig:pose_orbit}
    \vspace{-2mm}
\end{figure}

Training with SpatialVID improves pose accuracy at both spans and at every threshold (Tab.~\ref{tab:supp_orbit}). Fig.~\ref{fig:pose_orbit} shows example trajectories. The model trained on DL3DV only drifts on scenes with moving people, while the model trained with SpatialVID stays close to the ground truth. Accuracy at the strictest thresholds are low for both, showing that the large rotations in ORBIT remain challenging.
\begin{table}[h]
\centering
\setlength{\tabcolsep}{5pt}
\renewcommand{\arraystretch}{0.9}
\setlength{\aboverulesep}{1pt}
\setlength{\belowrulesep}{1pt}
\setlength{\abovecaptionskip}{3pt}
\resizebox{\linewidth}{!}{%
\begin{tabular}{cl c cccccc}
\toprule
$S$ & Training data & $N$ & RPA@3 $\uparrow$ & RPA@5 $\uparrow$ & RPA@10 $\uparrow$ & RPA@15 $\uparrow$ & RPA@30 $\uparrow$ & AUC@30 $\uparrow$ \\
\midrule
\multirow{2}{*}{64} & DL3DV & \multirow{2}{*}{330} & 5.5 & 13.6 & 33.7 & 48.1 & 71.6 & 42.5 \\
 & DL3DV + SpatialVID & & \textbf{7.5} & \textbf{17.3} & \textbf{39.2} & \textbf{55.9} & \textbf{76.7} & \textbf{47.9} \\
\midrule
\multirow{2}{*}{96} & DL3DV & \multirow{2}{*}{220} & 5.3 & 12.5 & 31.4 & 46.2 & 68.4 & 40.3 \\
 & DL3DV + SpatialVID & & \textbf{7.3} & \textbf{16.4} & \textbf{36.7} & \textbf{51.9} & \textbf{71.7} & \textbf{44.7} \\
\bottomrule
\end{tabular}%
}
\caption{Zero-shot camera pose accuracy on ORBIT (dynamic) dataset. $S$ is the frame span of each 10-view window, and $N$ is the number of windows.}
\label{tab:supp_orbit}
\end{table}
\vspace{-5mm}

\begin{figure}[!h]
    \centering
    \includegraphics[width=\linewidth]{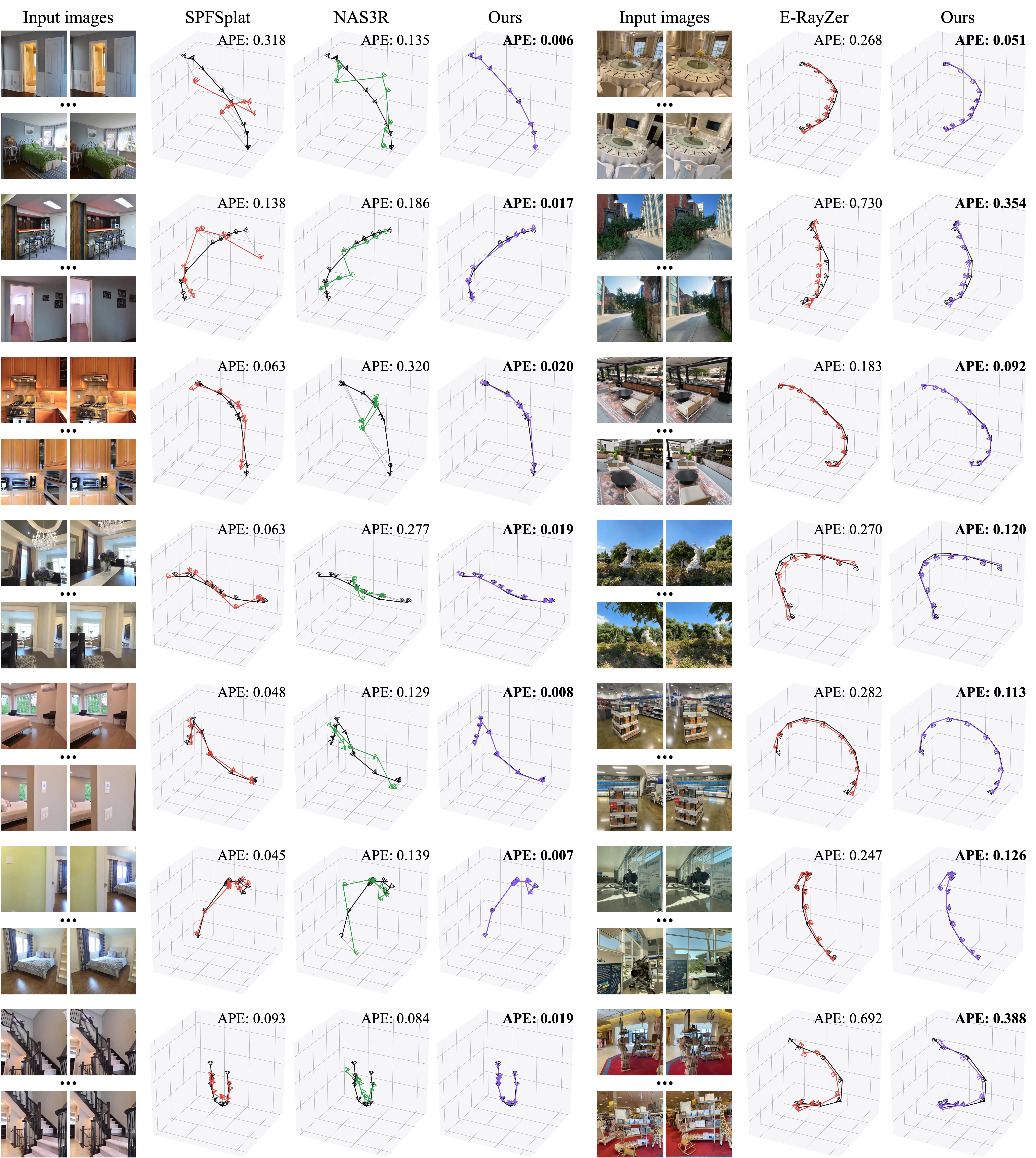}
    \vspace{-5mm}
    \caption{Additional qualitative comparison of recovered camera trajectories. In-domain camera poses from 10 unposed views on RE10K (left) and DL3DV (right). Predictions are shown in color and ground truth in black, aligned by a similarity transform.}
    \label{fig:sup_qual}
    \vspace{-2mm}
\end{figure}

\clearpage
\textbf{More qualitative results.} Fig.~\ref{fig:sup_qual} shows additional
camera trajectories recovered by each method.

\end{document}

%% file: math_commands.tex
\usepackage{amsmath,amsfonts,bm}

\def\eqref#1{equation~\ref{#1}}

\def\1{\bm{1}}

\DeclareMathAlphabet{\mathsfit}{\encodingdefault}{\sfdefault}{m}{sl}
\SetMathAlphabet{\mathsfit}{bold}{\encodingdefault}{\sfdefault}{bx}{n}

